\documentclass[]{clv3}
\usepackage{lineno,hyperref}
\usepackage{amsmath}
\usepackage{caption}
\usepackage{booktabs}
\usepackage{array}
\usepackage{makecell}
\usepackage{multirow}
\usepackage{adjustbox}

\usepackage{url}
\usepackage{float}
\usepackage[skip=10pt]{subcaption}
\usepackage{tikz}
\usepackage{CJKutf8}
\usetikzlibrary{bayesnet}
\usetikzlibrary{arrows}

\DeclareMathOperator{\E}{\mathbb{E}}
\usepackage{xcolor}
\definecolor{darkblue}{rgb}{0, 0, 0.5}
\hypersetup{colorlinks=true,citecolor=darkblue, linkcolor=darkblue, urlcolor=darkblue}

\issue{1}{1}{2016}

\dochead{Noun2Verb: Probabilistic frame semantics for word class conversion}

\runningtitle{Probabilistic frame semantics for word class conversion}

\runningauthor{Yu and Xu}

\begin{document}

\title{}

\author{Lei Yu\thanks{Department of Computer Science, University of Toronto. Email: jadeleiyu@cs.toronto.edu }}
\affil{University of Toronto}

\author{Yang Xu\thanks{Department of Computer Science, Cognitive Science Program,  University of Toronto. Vector Institute for Artificial Intelligence. Email: yangxu@cs.toronto.edu}}
\affil{University of Toronto}
\pageonefooter{
Action editor: Saif Mohammad; Submission received: 12 August 2021; Revised version received: 4 March 2022; Accepted for publication: 30 March 2022
}

\maketitle

\begin{abstract}
Humans can flexibly extend word usages across different grammatical classes, a phenomenon known as word class conversion. Noun-to-verb conversion, or denominal verb (e.g., \textit{to \underline{Google} a cheap flight}), is one of the most prevalent forms of word class conversion. However, existing natural language processing systems are impoverished in interpreting and generating novel denominal verb usages. Previous work has suggested that novel denominal verb usages are comprehensible if the listener can compute the intended meaning based on shared knowledge with the speaker. Here we explore a  computational formalism for this proposal couched in frame semantics. We present a formal  framework, {\it Noun2Verb}, that simulates the production and comprehension of novel denominal verb usages by modeling shared knowledge of speaker and listener in semantic frames. We evaluate an incremental set of probabilistic models that learn to interpret and generate novel denominal verb usages via paraphrasing. We show that a model where the speaker and listener cooperatively learn the joint distribution over semantic frame elements better explains the empirical denominal verb usages than state-of-the-art language models, evaluated against data from 1) contemporary English in both adult and child speech, 2) contemporary Mandarin Chinese, and 3) the historical development of English. Our work grounds word class conversion in probabilistic frame semantics and bridges the gap between natural language processing systems and humans in lexical creativity.
\end{abstract}

\section{Introduction}
Word class conversion refers to the extended use of a word across different grammatical classes without overt changes in word form. Noun-to-verb conversion, or denominal verb, is one of the most commonly observed forms of word class conversion. For instance, the expression {\it to Google a cheap flight} illustrates the innovative verb usage of {\it Google}, which is conventionally a noun denoting the web search engine or company. The extended verb use here signifies the action of ``searching information online''. Although denominal verbs have been studied extensively in linguistics as a phenomenon of lexical semantic innovation in adults and children and across different languages (e.g., ~\citealt{clark1979nouns,clark1982theyoung,vogel2011approaches,jespersen2013modern}), they have been largely under-explored in the existing literature of computational linguistics and their flexible nature presents key challenges to  natural language understanding and generation of innovative word usages. We present a formal computational account of noun-to-verb conversion couched in frame semantics. We show how our probabilistic framework yields sensible interpretation and generation of novel denominal verb usages that go beyond state-of-the-art language models in natural language processing.

\begin{figure}[h]
\subfloat[]{%
  \includegraphics[width=.95\columnwidth]{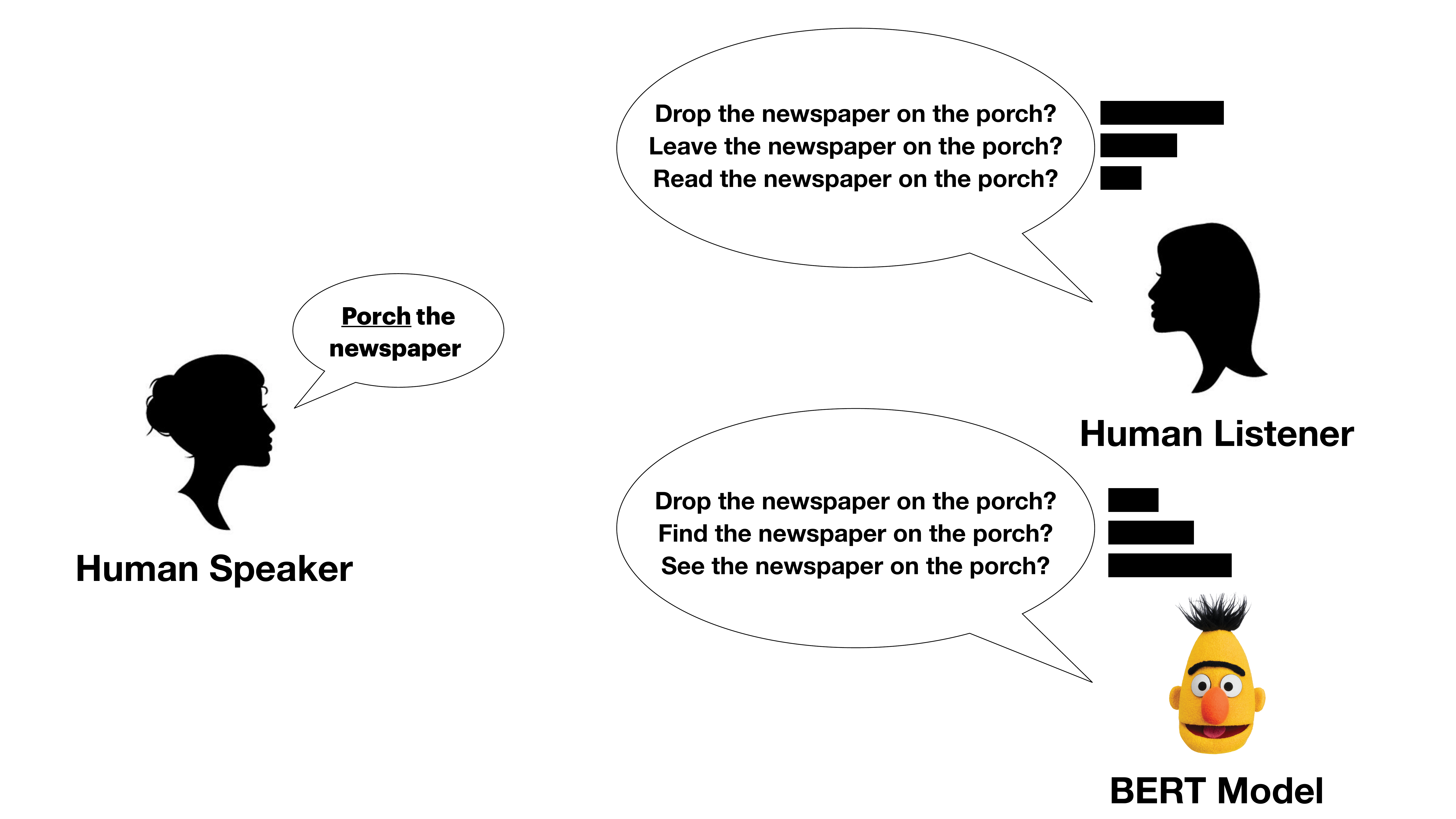}%
  \label{illustration_1}
}

\subfloat[]{%
  \includegraphics[clip,width=.95\columnwidth]{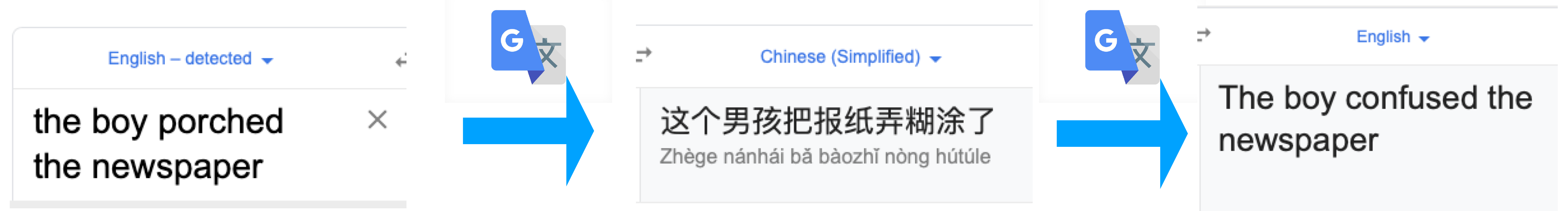}%
  \label{illustration_2}
}

\caption{Illustrations of the problem of noun-to-verb conversion, or denominal verb, in human language users and natural language processing systems. (a) Given a novel denominal usage of the noun {\it porch} uttered by the speaker, the listener  interprets the speaker's intended meaning correctly from context by choosing the most probable interpretation among a set of possible construals or paraphrases (bar length indicates probability of an interpretation). In comparison, the BERT language model assigns higher probabilities to inappropriate interpretations of the same denominal utterance. (b) The Google Translate system (evaluated in June 2021) incorrectly interprets \textit{to porch the newspaper} as \textit{to confuse the newspaper} when translating the query denominal utterance into Mandarin Chinese, which in turn leads to the erroneous back-translation into English.}

\end{figure}

Previous work has offered extensive empirical investigations into when noun-to-verb conversion occurs from the viewpoints of syntax \cite{hale1999bound}, semantics \cite{dirven1999conversion}, and pragmatics \cite{clark1979nouns}. In particular,  \citet{clark1979nouns} present one of the most comprehensive studies on this topic and describe ``the innovative denominal verb convention'' as a communicative scenario where the listener can readily comprehend the meaning of a novel denominal verb usage based on Grice's cooperative principles \cite{grice1975logic}. They suggest that the successful comprehension of a novel or previously unobserved denominal verb usage relies on the fact that the speaker denotes the kind of state, event, or process that s/he believes the listener can readily and uniquely compute on the basis of their mutual knowledge. They illustrate this idea with the classic example {\it the boy porched the newspaper} (see also Figure~\ref{illustration_1}). Upon hearing this utterance that features the novel denominal use of {\it porch}, the listener is expected to identify the scenario of a boy delivering the newspaper onto a porch, based on the shared world knowledge about the entities invoked by the utterance: the boy, the porch, and newspaper delivery systems. 

In contrast to human language users, existing natural language processing systems often fail to interpret (or generate) flexible denominal utterances in sensible ways. Figure~\ref{illustration_1} illustrates this problem in two established natural language processing systems. In Figure~\ref{illustration_1}, a state-of-the-art BERT language model assigned higher probabilities to two inappropriate paraphrases for the query phrase \textit{to porch the newspaper} over the more reasonable paraphrase \textit{to drop the newspaper on the porch}. In Figure~\ref{illustration_2}, the Google Translate system also failed to back-translate the same query denominal utterance from Mandarin Chinese to English. Specifically, this system  misinterpreted the denominal verb ``to porch'' with the translation ``to confuse'' in Mandarin Chinese, which led to the erroneous back-translation into English. These failed cases demonstrate the challenges toward natural language processing systems in interpreting flexible denominal verb usages, and they suggest that a principled computational methodology for supporting automated interpretation and generation of novel denominal verb usages may be warranted.



Work from cognitive linguistics, particularly  frame semantics, provides a starting point for tackling this problem from the view of structured meaning representation. Specifically, frame semantics theory asserts that humans understand word meaning by accessing a coherent mental structure of encyclopedic knowledge, or \textit{semantic frames}, that store a complex series of events, entities and scenarios along with a group of participants \cite{fillmore1967case}. Similar conceptual structures have also been discussed by researchers in artificial intelligence, cognitive psychology and linguistics, under the different terminologies of schema \cite{minsky1974framework, rumelhart1975notes}, script \cite{schank1972conceptual}, idealized cognitive model \cite{lakoff2008women, fauconnier1997mappings}, and qualia \cite{pustejovsky1991generative}. In the context of noun-to-verb conversion, frame semantics theory provides a principled foundation for characterizing human interpretation and generation of novel denominal verb usages. For example, the utterance ``the boy porched the newspaper'' may be construed as invoking a NEWSPAPER DELIVERY frame which involves both explicit \textit{frame elements} including the DELIVERER (the boy), the DELIVEREE (the newspaper) and the DESTINATION (the porch), as well as two latent elements that are left under-specified for the listener to infer: the main verb (also known as LEXICAL UNIT) that best paraphrases the action of the DELIVERER during the delivery event (e.g., {\it drop}), and the semantic relation between the DELIVERER and DELIVEREE (in this case can be described using a preposition ``on/onto''). Interpreting novel denominal usages, therefore, can be considered as a task of implicit semantic constituents inference which has  been explored in the paraphrasing of other types of compound expressions such as noun-noun compounds \cite{shwartz-dagan-2018-paraphrase, butnariu2009semeval}, adjective-noun pairing \cite{lapata2001corpus, boleda2013intensionality}, and logical metonymy \cite{lapata2003probabilistic}. 


The prevalence of denominal verb usages is not constrained to contemporary English. Apart from being observed in adult and child speech of different European languages \cite{clark1982theyoung, tribout2012verbal, mateu2001relational}, denominal verbs also commonly appear in more analytic languages (i.e., languages that rely primarily on helper words instead of morphological inflections to convey word relationships) such as Mandarin Chinese, where the absence of inflectional morphemes allows highly flexible shift from one word class to another \cite{wang2001dissertation,fang2000denom,si1996comparative}. From a historical point of view, many denominal verbs have emerged after the established usages of their parent nouns. For instance, according to the Oxford English Dictionary, the word {\it advocate} had been exclusively used as a noun denoting ``a person who recommends/supports something'' before 1500s. However, this word  grew a verb sense of ``to act as an advocate for something'' which later became popular so quickly that Benjamin Franklin in 1789 complained to Noah Webster about such an ``awkward and abominable'' denominal use \cite{franklin1789towebster}. It is therefore constructive to consider and evaluate a general formal framework for noun-to-verb conversion that supports denominal verb inference and generation across languages and over time.

\begin{figure}[t]
\centering
\begin{subfigure}[t]{0.85\linewidth}
   \includegraphics[width=1\linewidth]{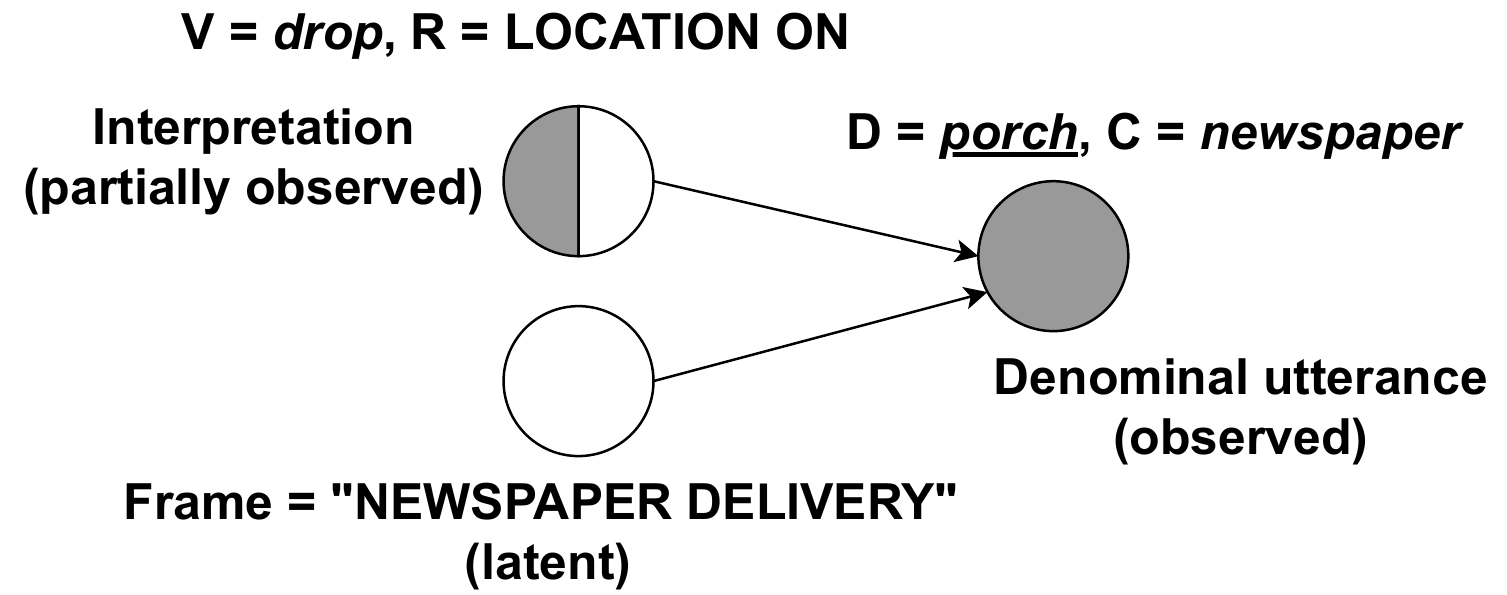}
   \caption{}
   \label{model} 
   \vspace*{3mm}
   
\end{subfigure}

\begin{subfigure}[t]{0.95\linewidth}
   \includegraphics[width=1\linewidth]{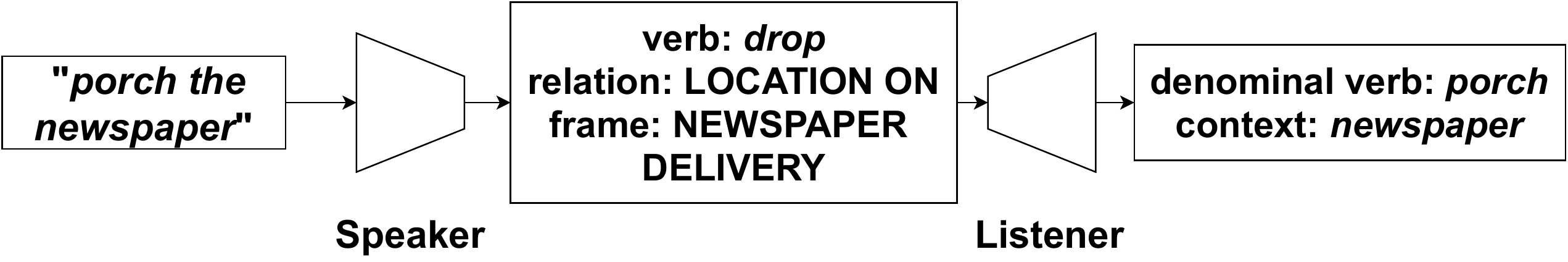}
   \label{reconstruction}
   \vspace*{3mm}
   \caption{}
\end{subfigure}
\caption{(a) The probabilistic graphical model of our {\it Noun2Verb} framework. (b) An illustration of the learning paradigm of \textit{Noun2Verb} based on the reconstruction process.}
\label{noun2verb-preview}
\end{figure}

In this work, we develop and evaluate a probabilistic framework, \textit{Noun2Verb}, to model noun-to-verb conversion couched in the tradition of frame semantics. Our work extends the previous study~\cite{yu2020nouns} which offers a probabilistic generative approach to model the meaning of denominal verb usages as a collection of frame elements. As illustrated in Figure ~\ref{noun2verb-preview}a, we use a probabilistic graphical model to capture the dependency over denominal utterances and their underlying frame elements, including 1) a partially observed interpretation of the denominal utterance consisting of a paraphrase verb and a semantic relation, and 2) a set of latent frame elements that further specify the underlying scenario. As shown in Figure ~\ref{noun2verb-preview}b, our framework maximizes the joint probability of the three types of variables via a communicative process between a listener module and a speaker module. These modules learn collaboratively to reconstruct a novel denominal utterance. In particular, the listener would first observe an utterance with novel denominal usages, and ``thinks out loud'' about its appropriate interpretation, which is then taken by the speaker as a clue to infer the actual denominal utterance. Intuitively, this process can succeed only if the listener interprets the denominal utterance correctly, and the speaker shares similar semantic frame knowledge with the listener. This learning scheme therefore operationalizes the mutual-knowledge-based communication proposed in \cite{clark1979nouns}. Moreover, the reconstruction process also allows the models to learn from denominal utterances without explicit interpretation in an unsupervised way.
To enable efficient learning, our framework draws on recent development from deep generative modeling \cite{Kingma2014AutoEncodingVB, kingma2014semi} and utilizes variational inference for training and learning with a minimal amount of labeled data. 

Our current study extends earlier work showing how this probabilistic generative model provides automated interpretation and generation of novel denominal verb usages in modern English~\cite{yu2020nouns}. We take a frame-semantic approach and compare three models of incremental complexity that range from a discriminative transformer-based model to a full generative model. We show that the  transformer-based model, despite its success in many natural language understanding tasks \cite{devlin2018bert}, is insufficient to capture the flexibility of denominal verbs and fails to productively generate novel denominal usages with relatively sparse training samples. We go beyond the previous work by a comprehensive evaluation of the framework with two additional sources of data: historical data of English noun-to-verb conversions and Mandarin denominal verb usages. Furthermore, we perform an in-depth analysis to interpret the learning outcomes of the generative model.

The remainder of this paper is organized as follows. We first provide an overview of the relevant literature. We then present our computational framework, \textit{Noun2Verb}, and specify the predictive tasks for model evaluation. We next present the datasets that we have collected and made publicly available for model learning and evaluation. We describe three case studies where we evaluate our framework rigorously on a wide range of data in contemporary English, Mandarin Chinese, and the historical development of English over the past two centuries. We finally provide detailed interpretations and discussion about the strengths and limitations of our framework and conclude.

\section{Related work}
\subsection{Computational studies on word class conversion}
Compared to the extensive empirical and theoretical research on word class conversion, very few studies have attempted to explore this problem from a computational perspective. One of the existing studies leverages recent advances in distributed word representations and deep contextualized language models to investigate the directionality of word class conversion. In particular, \citet{kisselew2016predicting} build a computational model to study the factors that may account for historical ordering between    noun-to-verb conversions and  verb-to-noun conversions in English. In that study, they train a logistic regression model using bag-of-words embeddings of lemmas attested in both nominal and verbal contexts to predict which word class (between noun and verb classes) might have emerged earlier in history. Their results suggest that denominal verbs usually have lower corpus frequencies than their parent noun counterparts, and nouns converted from verbs tend to have more semantically specific linguistic contexts. In a related recent study, \citet{li2020wordclass} perform a computational investigation on word class flexibility in 37 languages by using the BERT deep contextualized language model to quantify semantic shift between word classes. They find greater semantic variation when flexible lemmas (i.e., lemmas that have more than one grammatical class) are used in their dominant word class, supporting the view that word class flexibility is a directional process.

Differing from both of these studies, here we focus on modeling the process of noun-to-verb conversion as opposed to the directionality or typology of word class conversion across languages. 


\subsection{Frame semantics}

The computational framework we propose is grounded in  frame semantics which has a long tradition in linguistics and computational linguistics. According to \citet{fillmore2003background}, a semantic frame can potentially be evoked by a set of associated lexical units, which are often instantiated as the main predicate verbs in natural utterances. Each frame in the lexicon also enumerates several roles corresponding to facets of the scenario represented by the frame, where some roles can be omitted or null-instantiated and left under-specified for the listener to infer \cite{ruppenhofer2014null}. The problem of interpreting denominal verb usages can therefore be considered as inferring (the concepts evoked by) latent lexical unit(s) of the underlying semantic frame, which is itself related to the tasks of semantic frame identification \cite{hermann2014semantic} and semantic role labeling \cite{gildea2002automatic}. Given the limited available resources for labelled or fully annotated data, many existing studies have considered a generative and semi-supervised learning approach to combine annotated lexical databases such as FrameNet \cite{baker1998berkeley} and PropBank \cite{kingsbury2002treebank} with other unannotated linguistic corpora. For instance, the SEMAFOR parser presented by \citet{das2014frame} is a latent variable model that learns to maximize the conditional probabilities of labeled semantic roles in FrameNet, and supports lexical expansion to unseen lexical units via the graph-based semi-supervised learning technique \cite{bengio200611}. In a separate work, \citet{thompson2003generative} learn a generative Hidden Markov Model using the labeled sentences in FrameNet and show that the resulting model is able to infer null-instantiated semantic roles in unobserved utterances (e.g., inferring that a ``driver'' role is missing given the sentence {\it The ore was boated down the river}). 

Our framework builds on these existing studies by formulating noun-to-verb conversion as probabilistic inference of latent semantic frame constituents, and we suggest how a semi-supervised generative learning approach offers data efficiency and effective generalizations on the interpretation and generation of novel denominal verb usages that do not appear in the training data.

\subsection {Models of compound paraphrasing}

Our study also relates to a recent line of research on  compound understanding. Many problems concerning the understanding of compounds require the inference of latent semantic constituents from linguistic context. For example, \citet{nakov2006using} suggest that the semantics of a noun-noun compound can be expressed as multiple prepositional and verbal paraphrases (e.g., \textit{apple cake}
can be interpreted as \textit{cake made of/contains apples}). Later work develops both supervised and unsupervised learning approaches to tackling noun-compound paraphrasing \cite{van2013melodi, xavier2014boosting}. In particular, \citet{shwartz-dagan-2018-paraphrase} propose a semi-supervised learning framework for inferring the latent semantic relations of noun-noun compounds. They represent compounds and their paraphrases in a distributed semantic space parameterized by a biLSTM \cite{graves2005framewise} encoder. When paraphrases are not available, the missing components are replaced by the corresponding hidden representations yielded by the encoder. \citet{shwartz-dagan-2018-paraphrase} show good generalizability of their model on unobserved examples. We show that our framework generalizes well on novel denominal utterances due to a semi-supervised learning approach in a distributed semantic space, and further, the proposed framework can learn interpretation (listener) and generation (speaker) model simultaneously via generative modeling.

Previous linguistic studies also suggest that the lexical information in converted denominal verbs can be inferred from the listeners' knowledge about the intended referent of nominal bases \cite{baeskow2006reflections}. It is therefore natural to connect noun-to-verb conversion to the linguistic phenomenon of logical metonymy, where language users   need to infer missing predicates from certain syntactic constructions (e.g., \textit{an easy book} means \textit{a book that is easy to read}) \cite{pustejovsky1991generative}. Following this line of thought, \citet{lapata2003probabilistic} propose a probabilistic model that can rank interpretations of given metonymical compounds by searching in a large corpus for their paraphrases, which are identified by exploiting the consistent correspondences between surface syntactic cues and meaning. We  apply similar methods to extract candidate paraphrases of denominal utterances to construct our learning or training dataset, and we show that this frequency-based ranking scheme aligns reliably with human feasibility judgement of interpretations for denominal verb usages.  

\subsection{Deep generative models for natural language processing}

The recent surge of deep generative models has led to the development of several flexible language generation systems, such as variational autoencoders (VAEs) \cite{bowman-etal-2016-generating, bao-etal-2019-generating, fang2019implicit} and generative adversarial networks (GANs) \cite{subramanian-etal-2017-adversarial, press2017language, lin2017adversarial}. Our Noun2Verb framework builds on the architecture of semi-supervised VAE proposed by \citet{kingma2014semi}, where an interpretation/listener module and a generation/speaker module jointly learn a probability distribution over all denominal utterances and any of their available paraphrases. One advantage of VAEs is the ability to encode through their latent variables certain aspects of semantic information (e.g., writing style, topic, or high-level syntactic features), and to generate proper samples from the learned hidden semantic space via ancestral sampling. We show in our model analysis that the learned latent variables in our framework indeed capture the variation in both syntactic structures and semantic frame information of target denominal utterances and their paraphrases.

\subsection{Deep contextualized language models}
For a sequence of natural language tokens, deep contextualized models compute a sequence of context-sensitive embeddings for each token. Many state-of-the-art natural language processing models are built upon stacked layers of a neural module called the Transformer \cite{vaswani2017attention}, such as BERT \cite{devlin2018bert}, GPT-2 \cite{radford2019language}, RoBERTa \cite{liu2019roberta}, and BART \cite{lewis-etal-2020-bart}. These large neural network models are often pre-trained on predicting missing tokens given contextual information within a sentence. The models are then fine-tuned on learning examples of a series of downstream tasks including language generation tasks such as summarization, and natural language understanding tasks such as recognizing textual entailment. A common issue of most current transformer-based models is that many of their successful applications tend to rely on extensive fine-tuning on adopted benchmarks with (sometimes hundreds of) thousands of examples. For tasks where large-scale annotations of learning examples are infeasible, or where the target linguistic data are severely under-represented in standard pre-training resources, transformer models often yield much worse performance \cite{croce-etal-2020-gan}. 

In our work, we consider a BERT-based language generation model as a competitive baseline, and we demonstrate that this pre-trained language model is insufficient to capture the flexibility of noun-to-verb conversions, particularly when ground-truth paraphrases for a denominal utterance are highly uncertain.

\section{Computational framework}
We formalize noun-to-verb conversion as a dual problem of comprehension and production and formulate this problem under a frame semantics perspective. We present three incremental probabilistic models under differing assumptions about the computational mechanisms of noun-to-verb conversion.

\subsection{Noun-to-verb conversion as probabilistic inference}

We consider noun-to-verb conversion as communication between a listener and a speaker over an utterance that includes a novel denominal verb usage. Our framework focuses on modeling the knowledge and dynamics that enable 1) a listener module to properly interpret the meaning of a novel denominal verb usage (or zero-shot inference) by paraphrasing, and 2) a speaker module to produce a novel denominal usage given an interpretation.

\begin{figure}[t]
\centering
\includegraphics[width=13cm]{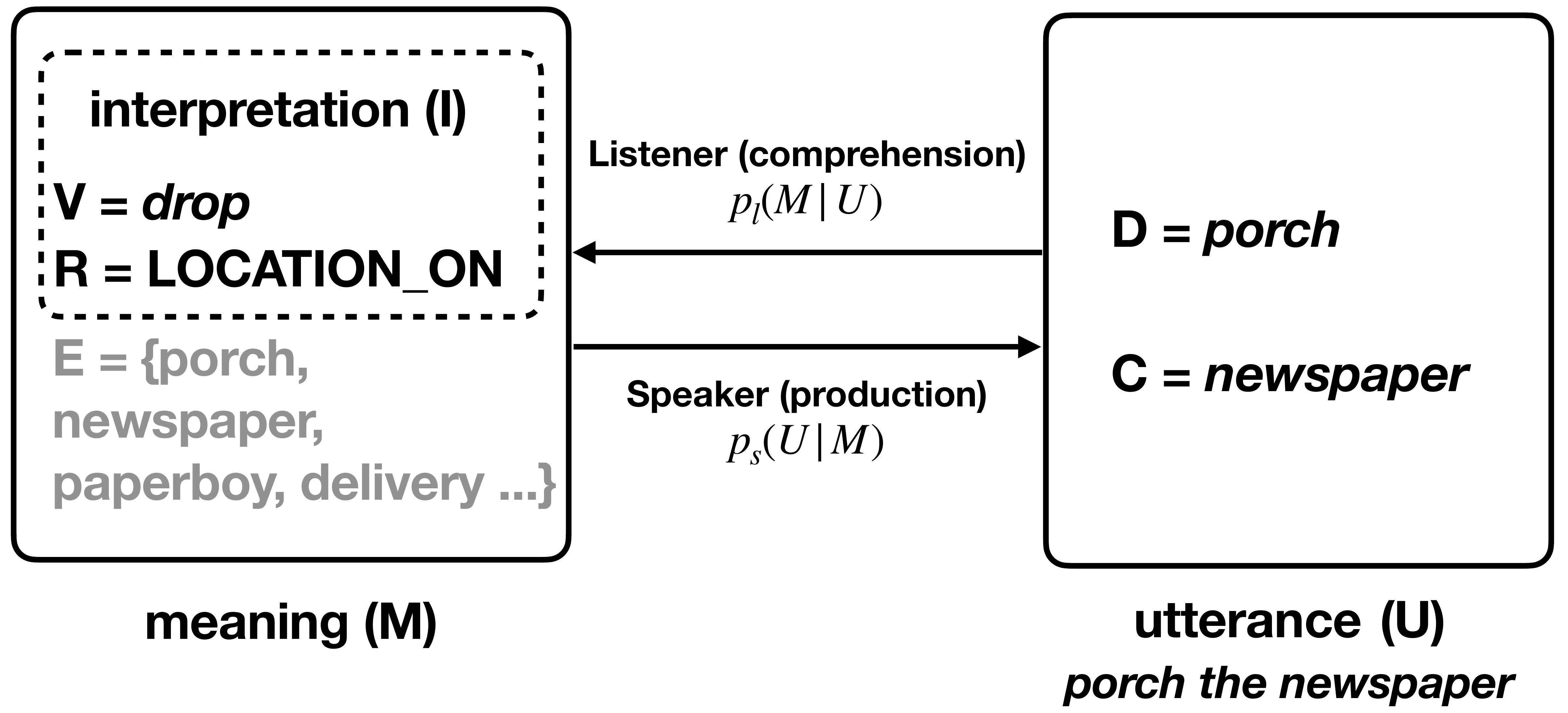}
\caption{An illustration of the \textit{Noun2Verb} framework. The speaker produces an utterance of a denominal verb usage from its production likelihood $\text{p}_{s}$. The listener interprets the meaning of the utterance by paraphrasing via its comprehension likelihood $\text{p}_{l}$. $U = (D,C)$ is the denominal utterance, where $D$ is the target denominal verb, and $C$ its object context; $M$ is the meaning of $U$, where $V$ is the paraphrased verb, $R$ is the semantic relation, and $E$ denotes a set of latent frame elements.}
\label{n2v_framework}
\end{figure}

Figure~\ref{n2v_framework} illustrates our framework. Here the speaker generates an utterance $U = (D,C)$ that consists of an innovative usage of denominal verb $D$  (e.g., \textit{porch}) and its \textit{context} $C$. As an initial step, we consider the simple case where $C$ is a single word that serves as the direct object of $D$ (e.g., \textit{newspaper} as the context for \textit{porch}).

Given this utterance, the listener interprets its meaning $M$ which we operationalize as three key components: 1) a paraphrase verb $V$ (e.g., {\it drop}) for the target denominal verb; 2) a \textit{semantic relation}  $R$ following \citet{clark1979nouns} that specifies the relation between the paraphrase verb and the context (e.g., an on-type location, signifying the fact that newspaper is dropped onto the porch); 3) a set of frame elements $E$ following the frame semantics tradition, which we elaborate below. The paraphrase verb $V$ is an established verb that best describes the action denoted by the denominal verb $D$. It serves as the \textit{lexical unit} that invokes the underlying semantic frame of $D$. The semantic relation $R$, according to empirical studies in \citet{clark1979nouns}, reflects how the novel sense of a denominal verb is extended from its parent noun, and falls systematically into eight main types (see a summary in Table~\ref{rel_table}). Within each relation type, there is a set of words (most of which are prepositions) that signify such a relation, along with a template paraphrase for denominal usages of this type. For instance, denominal usages of the form ``to $<$denominal verb$>$ the $<$context$>$'' (e.g., ``to porch the newspaper'') where $D$ comes from the relation type LOCATION ON can usually be paraphrased as ``to $<$paraphrase verb$>$ the $<$context$>$ onto/into/to the $<$denominal verb$>$'' (e.g., ``to drop the newspaper onto the porch''). Under a semantic frame invoked by $V$, the listener would simultaneously infer  frame elements $E$ that may be involved in the scenario expressed by the target utterance $U$---such inference captures not only participants that are explicitly specified by the denominal utterance, but also the residual contextual knowledge shared between the speaker and the listener that is not captured in variables $V$ and $R$. In particular, {\it porch the newspaper} may invoke a DELIVERY frame, where one can identify that the element of DELIVERY is {\it the newspaper}, the destination is {\it the porch}, and infer that a reasonable choice of the DELIVERER role can be {\it the postman} or {\it the paperboy}. We denote $I = (V,R)$ as an interpretation for a target utterance $U$, while we specify frame elements $E$ as latent variables (i.e., implicit knowledge) to be inferred by the models.

\begin{table}[bt]
\caption{Major types of semantic relation described in \cite{clark1979nouns} that explain common denominal verb usages in English. Each semantic relation is specified by a set of relational words (mostly prepositions), and a syntactic schema that serves as the template for paraphrasing query denominal verb usages under a relation type.}
\centering
\begin{tabular}{|p{2.7cm}|p{2.7cm}|p{2.7cm}|p{2.7cm}| }
\hline
\thead{Relation type} & \thead{Relational words} & \thead{Denominal usage} & \thead{Template paraphrase} \\
\hline
LOCATUM ON & {\it on, onto, in, into, to, at} & \underline{carpet} the floor & put the carpet on the floor \\
\hline
LOCATUM OUT & {\it out (of), from, of} & \underline{shell} the peanuts & remove the shell from the peanuts \\
\hline
LOCATION IN & {\it on, onto, in, into, to, at} & \underline{porch} the newspaper & drop the newspaper on the porch \\
\hline
LOCATION OUT & {\it out (of), from, of} & \underline{mine} the gold & dig the gold out of the mine  \\
\hline
DURATION & {\it during} & \underline{weekend} at the cabin & stay in the cabin during the weekend \\
\hline
AGENT & {\it as, like} & \underline{referee} the game & watch the game as a referee \\
\hline
GOAL & {\it become, look like, to be, into} & \underline{orphan} the children & make the children become orphans \\
\hline
INSTRUMENT & {\it with, by, using, via, through} & \underline{bike} to school & go to school by bike \\
\hline  
\end{tabular}

\label{rel_table}
\end{table}

Our formulation drawing on semantic relations is motivated by existing cross-linguistic studies of denominal verbs. For example, Clark found that semantic relation types in Table~\ref{rel_table} apply to many innovative denominal usages coined by children speaking English, French, and German \cite{clark1982theyoung}. A more recent comparative study of denominal verbs in English and Mandarin Chinese also found that these major semantic relations can explain many Chinese denominal verb usages \cite{Bai2014DenominalVI}. The modeling framework we present here can automatically learn these semantic relations and latent frame elements from data, and importantly it can generalize to interpret and generate novel denominal usages across different languages and over time.

With the core components defined, we now formally cast noun-to-verb conversion as two related probabilistic inference problems. The listener module tackles the \textit{comprehension problem} where given an utterance $U$, it samples appropriate paraphrases to interpret its meaning $M = (I, E) = (V, R, E)$ under the comprehension model $\text{p}_{l}(M|U)$. The speaker module tackles the inverse \textit{production problem} by producing a (novel) denominal usage $U$ given an intended meaning $M$, under the production model $\text{p}_{s}(U|M)$. 

We postulate  that  mutually shared knowledge, when modeled as semantic frame information, should be key to successful communication for innovative denominal usages. To verify this view, we describe and examine three incremental probabilistic models under our framework dubbed \textit{Noun2Verb}.

\subsection{Model classes}

\begin{figure}[t]
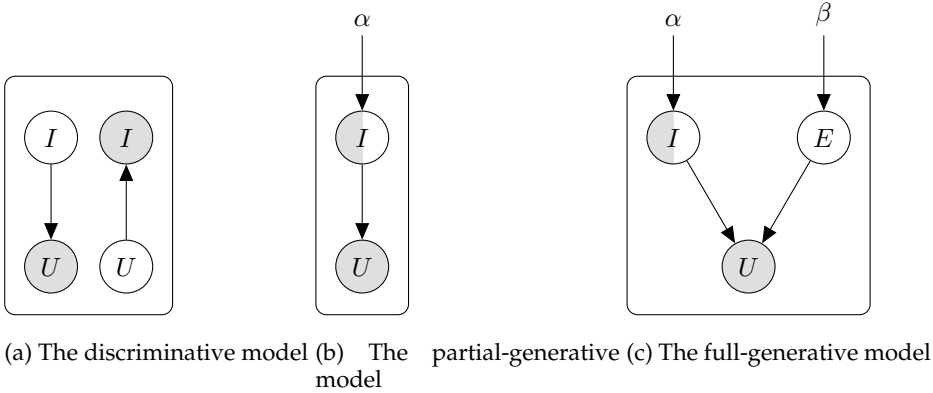

   \begin{subfigure}[t]{.30\linewidth}
  \tikz{
 \node[obs] (x1) {$U$};%
 \node[latent,above=of x1,xshift=0cm] (y1) {$I$}; %

 \node[latent, xshift=1cm] (x2) {$U$};%
 \node[obs,above=of x2,xshift=0cm] (y2) {$I$}; %
 \plate [inner sep=.25cm,yshift=.2cm] {plate1} {(x1)(y1)(x2)(y2)} {}; %
 \edge {y1} {x1}  
 \edge {x2} {y2}  
 }
 \caption{The discriminative model}
  \end{subfigure}
   \begin{subfigure}[t]{.30\linewidth}
  \tikz{
 \node[obs] (U) {$U$};%
 \node[latent,above=of U,xshift=0cm,path picture={\fill[gray!25] (path picture bounding box.south) rectangle (path picture bounding box.north west);}] (I) {$I$}; %
 \node[above, above= of I](alpha) {$\alpha$};%

 \plate [inner sep=.25cm,yshift=.2cm] {plate1} {(U)(I)} {}; %
 \edge {I} {U} 
 \edge{alpha}{I}
 }
 \caption{The partial-generative model}
  \end{subfigure}
  \begin{subfigure}[t]{.30\linewidth}
  \tikz{
 \node[obs] (U) {$U$};%
 \node[latent,above=of U,xshift=-1cm,path picture={\fill[gray!25] (path picture bounding box.south) rectangle (path picture bounding box.north west);}] (I) {$I$}; %
 \node[latent,above=of U,xshift=1cm] (E) {$E$}; %
 \node[above, above= of I](alpha) {$\alpha$};%
 \node[above, above= of E](beta) {$\beta$};%

 \plate [inner sep=.25cm,yshift=.2cm] {plate1} {(U)(I)(E)} {}; %
 \edge {E,I} {U}  
 \edge{alpha}{I}
 \edge{beta}{E}
 
 }
 \caption{The full-generative model}
  \end{subfigure}
  \caption{A graphical illustration of the three probabilistic models under the proposed framework. $U$, $I$, and $E$ stand for the variables of (u)tterance that contains a denominal verb usage, (i)ntended meaning of the utterance, and (e)lements of the semantic frame invoked by the utterance, respectively. $\alpha$ and $\beta$ represent the hyperparameters for the prior distributions of the variables. Shaded, half-shaded, and unshaded circles represent latent variables, semi-latent variables, and  observables.}
  \label{PGM}
\end{figure}

We present three probabilistic models (see illustrations in Figure~\ref{PGM}) that make different assumptions about the computational mechanisms of noun-to-verb conversion. First, we describe a \textit{discriminative model} that assumes neither any interactive dynamics between the speaker and the listener (i.e., no collaboration) nor any knowledge of semantic frame elements. We implement this model using a state-of-the-art contextualized language model from natural language processing. To our knowledge, there exists no specific and scalable model of denominal verbs, and given the general-purpose nature of contextual language models, we consider it as a strong competitive baseline model. Next, we describe a \textit{partial generative model} that enables listener-speaker collaboration via knowledge sharing but without any representation of semantic frame elements. Finally, we describe a \textit{full generative model} that incorporates both listener-speaker collaboration and semantic frame elements.

\subsubsection{ Discriminative model}
The discriminative model consists of two sub-modules that learn separately without any collaboration or information sharing; hence it is insensitive to frame elements $E$ in its meaning representation. As illustrated in Figure~\ref{PGM}a, the listener module receives a denominal utterance, and produces a paraphrase of that utterance by sampling an interpretation from the conditional distribution $\text{p}_{l}(I=(V,R)|U=(D,C))$. The speaker module reverses the listener module by generating a denominal utterance from the conditional distribution $\text{p}_{s}(U=(D,C)|I=(V,R))$. During learning, we present the model with a supervised set (i.e., fully labelled data) of denominal utterances $X_\text{s} = \{(U^{(i)}, I^{(i)})\}_{i=1}^{M}$. Each of such data points is paired with a human-annotated ground-truth paraphrase verb and semantic relation (i.e., the interpretation for a query denominal usage). We optimize the speaker-listener modules by minimizing the standard negative log-likelihood classification loss for both modules independently:
\begin{align}
    \mathcal{S} &= \mathcal{S}_l + \mathcal{S}_s \\
    \mathcal{S}_l &= -\sum_{(U^{(i)}, I^{(i)}) \in X_\text{s}} \log \text{p}_{l}(I^{(i)}|U^{(i)}; \Theta_l) \\
    \mathcal{S}_s &= -\sum_{(U^{(i)}, I^{(i)}) \in X_\text{s}} \log \text{p}_{s}(U^{(i)}|I^{(i)}; \Theta_s))
\end{align}
Here $\Theta_s$ and $\Theta_l$ denote the parameters under the speaker and the listener modules, respectively. This discriminative learner bears resemblance to compound phrase understanding systems, where  classification models are trained to predict implicit semantic relations that hold between phrase constituents \cite{shwartz-dagan-2018-paraphrase}.

We consider the state-of-the-art language model BERT \cite{devlin2018bert} to parameterize the listener and speaker distributions  (see Appendix A for a detailed description of the BERT model architecture). However, as we will demonstrate empirically, despite the incorporation of such a powerful neural language model with a rich knowledge base, this discriminative baseline model is insufficient to simulate word class conversion in sensible ways, mostly due to its limitations in capturing the flexibility and uncertainty involved in natural denominal usages. For instance, both ``drop the newspaper on the porch'' and ``leave the newspaper on the porch'' can be considered good interpretations for the query denominal usage {\it porch the newspaper}, but systems like BERT as shown later tend to idiosyncratically favor a very restricted set of construals and cannot account for the fine-grained distribution of human interpretations for denominal usages. Furthermore, the speaker and listener modules in the discriminative model do not share mutual knowledge by jointly encoding the same probability distribution over denominal utterances and their interpretations; that is, $\text{p}_{l}(I|U)$ and $\text{p}_{s}(U|I)$ do not necessarily induce the same joint distribution $\text{p}(U,I)$. We therefore turn to a more cognitively viable generative model, by incorporating the interaction between the listener and speaker modules to encourage agreement on the utterance-meaning distributions --- a prerequisite for successful communication with innovative denominal usages~\cite{clark1979nouns}. 

\subsubsection{Partial generative model}
The partial generative model, illustrated in Figure~\ref{PGM}b, defines a generative process of how a speaker might produce a novel denominal usage. We first sample an interpretation by drawing $I$ from a categorical prior distribution $\text{p}_0(I|\bf{\alpha})$ parametrized by $\alpha$. We then feed this interpretation to the speaker module so as to sample a novel denominal utterance via $\text{p}_\text{s}(U|I)$. This setup enforces a joint utterance-interpretation distribution $\text{p}_\text{s}(U,I|\alpha) = \text{p}_0(I|\alpha)\text{p}_\text{s}(U|I)$, which allows us to operationalize the idea of shared mutual knowledge by encouraging the listener to be consistent with the speaker when interpreting novel denominal usages.

Formally, we learn the listener's likelihood $\text{p}_l(I|U)$ as a good approximation for the speaker's distribution $\text{p}_\text{s}(I|U)$ over interpretations:
\begin{align}
    \text{p}_\text{l}(I|U) &\approx \text{p}_\text{s}(I|U)
\end{align}
We parametrize model distributions $\text{p}_\text{l}$ and $\text{p}_\text{s}$ via feed-forward neural networks (see Appendix A for a detailed model description). 
One advantage of this generative approach is that it supports learning with sparse labelled data. In particular, this model can learn from a handful of labelled data and an unlabelled, or unsupervised set $X_\text{u}\{(U^{(i)})\}_{i=1}^{N}$, where each denominal verb usage has no human annotation (in terms of its meaning). 

To learn this model, we apply an optimization technique known as variational inference, commonly used for generating data with highly complex structures including images \cite{narayanaswamy2017learning} and text \cite{semeniuta2017hybrid}, to train the two modules simultaneously. Let $\Theta$ again denote the set of all parameters in the model; we optimize $\Theta$ by minimizing the following evidence lower bound (ELBO) loss function:
\begin{align}
    \mathcal{U} = \sum_{U^{(i)}\in X_\text{u}} \E_{I\sim \text{p}_\text{l}}[\log \text{p}_\text{s}(U|I)] - D[\text{p}_\text{l}(I|U)||\text{p}_0(I|\alpha)]
\end{align}
Here $\E_{I\sim p_\text{l}}(\cdot)$ refers to taking the expectation by sampling interpretation $I$ from the listener's conditional likelihood $p_\text{l}(I|U)$, and $D(\cdot||\cdot)$ denotes the Kullback-Leibler (KL) divergence between two probability distributions. This learning scheme does not require any labelled interpretation $I$. Instead, the two modules learn collaboratively by seeking to reconstruct a denominal verb utterance: the first term $\E_{I\sim p_\text{l}}[\log p_\text{s}(U|I)]$ of $\mathcal{U}$ describes a scenario where the listener first observes a $U$ and ``thinks out loud'' about its interpretation $I$, which is then taken by the speaker (who is hidden from the utterance) as a clue to infer the actual utterance. Intuitively, if the listener understands $U$ reasonably, and provided that the speaker shares a similar utterance-interpretation mapping with the listener, the reconstruction is more likely to succeed, and existing theoretical analyses validate this idea (see, for example, \cite{rigollet2007generalization}, for detailed discussion). It can be shown that minimizing $\mathcal{U}$ is equivalent to maximizing the joint log-likelihood of all denominal utterances in the unsupervised set, while simultaneously finding a listener's likelihood $p_\text{l}(I|U)$ that best approximates the speaker's posterior $p_\text{s}(I|U)$. We provide the proof of this equivalence in Appendix B for interested readers.

Apart from the above unsupervised learning procedure, we can also train the two modules separately on the labelled, supervised set $X_s$ just as we learn in the discriminative model. The overall learning objective $\mathcal{L}$, therefore, consists of minimizing jointly a supervised loss term and an unsupervised one, which can be operationalized through the paradigm of semi-supervised learning:
\begin{align}
    \mathcal{L} = \mathcal{U} + \lambda \mathcal{S}
\end{align}
Here $\mathcal{S}$, $\mathcal{U}$ are the two losses defined in Equations (1) and (5), and $\lambda$ is a hyperparameter controlling the relative weighting of the supervised and unsupervised data.
Training the partial generative model (as well as the full generative model described next) is algorithmically equivalent to learning a semi-supervised variational autoencoder proposed by \citet{kingma2014semi}.


\subsubsection{Full generative model}
Similar to the partial model, the full generative model illustrated in Figure~\ref{PGM}c also defines a generative process from meaning $M$ to utterance $D$, except that the semantic frame elements $E$ are incorporated as a latent variable: $\text{p}_s(U,I|\alpha,\beta) = \text{p}_0(I|\alpha)\text{p}_0(E|\beta)\text{p}_s(U|I)$, where $\alpha, \beta$ are hyperparameters that define the categorical priors of $I$ and $E$ respectively. In this model, both the interpretation $I$ and the semantic frame $E$ give rise to a denominal utterance. Intuitively, the introduction of frame elements helps the model to further distinguish denominal utterances of similar interpretations but distinct intended referents. For example, both of the denominal utterances 1) {\it carpet the floor} and 2) {\it blanket the bed} can be paraphrased by the same coarse, semantic-relation template ``to put A on (the top of) some B'', but their actual contexts are quite different. The frame element $E$ is expected to capture such fine-grained variation in meaning by learning the residual contextual information under-specified by $V$ and $R$. Similar to the partial generative model, we still expect an agreement between the posteriors of meaning $\text{p}_s(M|U)$ and $\text{p}_l(M|U)$, but here we use the full representation of $M=(I,E)$ by taking frame elements into consideration:
\begin{align}
    \text{p}_\text{l}(I,E|U) &\approx \text{p}_\text{s}(I,E|U)
\end{align}
The listener and speaker distributions here are parametrized by neural network encoders. During learning, the model can also be trained via a mixture of 1) reconstruction of unlabeled denominal utterance, and 2) inference and generation of labeled denominal usages with ground-truth paraphrases.
The unsupervised learning stage is also conducted through variational inference with an ELBO loss function similar to the partial model:
\begin{align}
    \mathcal{U} = \sum_{U^{(i)}\in X_\text{u}} \E_{(I,E)\sim \text{p}_\text{l}}[\log \text{p}_\text{s}(U|I,E)] - D[\text{p}_\text{l}(I,E|U)||\text{p}_0(I|\alpha)\text{p}_0(E|\beta)]
\end{align}
whereas the supervised learning loss is identical to $L$ in Equation (1), and the overall semi-supervised loss shares the same form as specified in Equation (4).

\subsection{Specification of predictive tasks}
We consider our models in two  predictive tasks: 1) in the {\it comprehension task}, the listener module of the model takes an utterance containing a novel query denominal usage $U$ and provides an interpretation of its meaning through sampling from $\text{p}_l(I|U)$ it defines; 2) in the {\it production task}, the speaker module conversely generates a novel denominal usage $U$ from its $\text{p}_s(U|I)$ based on a query meaning specified in $I$. For the full generative model, since $U$ depends on both interpretations and frame elements, we apply a Monte Carlo approach to approximate $p_s(U|I)$ and $p_l(I|U)$ by first drawing a set of frame elements $E^{(k)}$ from model priors, and then taking the average over the production probabilities $\text{p}_s(U|I,E^{(k)})$ induced by sampled elements:

\begin{align}
\text{p}_l(I|U)   &\approx \sum_{E^{(k)} \sim \text{p}_0(E|\alpha)} \text{p}_l(I,E^{(k)}|U) \\
\text{p}_s(U|I) &\approx \sum_{E^{(k)} \sim \text{p}_0(E|\alpha)} \text{p}_s(U|I,E^{(k)})
\end{align} 

For evaluation against historical data, we incrementally predict the denominal usages $U^{(t+\Delta)}$ of a target noun $D$ emerged at future time $t+\Delta$, given its established noun usages up to time $t$ -- for instance, we expect the model to infer whether the noun ``phone'' can grow out a verb sense given its nominal usage before 1880s. We formalize this temporal prediction problem by assuming that an appropriate denominal usage generated by the speaker should be acceptable to the language community in the future. We thus extend the synchronic production task to make diachronic prediction. In particular, the speaker module takes the predicate verbs and semantic relations associated with the target noun $D$ at time $t$ as interpretation $I^{(t)}$, and sample a denominal usage $\hat{U}^{(t)} \sim \text{p}_s(U|I^{(t)})$ as model prediction for denominal usages into the future times $t+\Delta$:
\begin{align}
\text{Pr}(U^{(t+\Delta)}|I^{(t)}) = \text{p}_s(U^{(t)}|I^{(t)})
\end{align} 
The full generative model $\text{p}_s(U^{(t)}|I^{(t)})$ is again approximated by the Monte Carlo sampling approach in Equation (8).

\section{Data}    
To evaluate our framework comprehensively against natural denominal verb usages, we collected three datasets: 1) Denominal verbs from adults and children speaking contemporary English extracted from the literature (DENOM-ENG); 2) 
Denominal verbs in contemporary Mandarin Chinese extracted from the literature (DENOM-CHN); 3) Denominal verbs extracted from historical English corpora (DENOM-HIST). Each dataset consists of a supervised set $X_s$ of denominal usages with interpretations, and an unsupervised set $X_u$ of unannotated denominal usages. We also collected a set of synchronic denominal verb usages with ground-truth paraphrases annotated via online crowdsourcing (DENOM-AMT) for model evaluation.\footnote{Data and code for our analyses are available at the following repository:  \url{https://github.com/jadeleiyu/noun2verb}.} The experimental protocol of this work has been approved by the Research Ethics Boards at the University of Toronto (REB \# 00036310). A total amount of 1,304 US dollars were paid to human annotators for about 13,000 responses. Every annotator received an estimated hourly payment that is higher than the minimum wage requirement in his/her registered country \footnote{The average payment in our task is 33.6 USD per hour, which is above the minimum wage requirements of the registered countries of all involved participants (from Canada, People's Republic of China, United Kingdom, and United States)}.


\subsection{Denominal verb usages from English-speaking adults and children (DENOM-ENG)}
\citet{clark1979nouns} provide a large list of denominal verb utterances (i.e., a denominal verb with its context word) from English adults, and \citet{clark1982theyoung} also reports a set of novel denominal uses produced by English-speaking children under age 7. Although all of these denominal utterances are labelled with their ground-truth relation types $R$, none of them has ground-truth paraphrase verb(s) $V$ available. To obtain natural interpretations of denominal meaning (for constructing the supervised set for model learning), we searched for the top 3 verbs that co-occur most frequently with each denominal utterance using the paraphrase templates specified in Table~\ref{rel_table} (and we validated these searched results using crowdsourcing described later). We performed these searches in the large-scale comprehensive iWeb 2015 corpus (\url{https://corpus.byu.edu/iweb/}), specifically through the Sketch Engine online corpus tool (\url{https://www.sketchengine.eu}) and its built-in regular-expression queries -- for example, a denominal utterance ``to carpet the floor'' with a ``LOCATUM ON'' relation type would have a paraphrase utterance template ``to \_\_ the carpet on/onto the floor'', where ``\_\_'' is filled by a verb. We obtained 786 annotated denominal utterances from adult data, and 32 annotated examples from children.   

While a small portion of denominal utterances has explicit human-annotated paraphrases, a greater proportion does not have such information. We expect our models to be able to interpret novel denominal verb usages by generalizing from the small set of annotated data and also learning from the large set of unlabelled data. For example, if the model is told that ``send the resume via email'' is the correct paraphrase for {\it email the resume}, then on hearing a similar utterance like {\it mail the package}, it should generalize and infer that utterance has something to do with the transportation frame (as in the case with {\it mail}). To facilitate such ``frame borrowing'' learning, we obtained a set of novel denominal usages by replacing the denominal verb $D$ of each $U$ described previously with a semantically related noun (e.g., {\it mail the letter} $\rightarrow$ {\it email the letter}). We took the taxonomy from WordNet (\url{https://wordnet.princeton.edu/}) and extracted all synonyms of each denominal verb $D$ from the same synset as substitutes. This yielded 1,129 novel utterances examples for unsupervised learning.

\subsection{Denominal verb usages in Mandarin Chinese (DENOM-CHN)}
Similar to the case of English, noun-to-verb conversion has been extensively investigated in Mandarin Chinese. In particular, \citet{Bai2014DenominalVI} performed a comparative study of denominal verbs in English and Mandarin Chinese by collecting over 200 examples of noun-to-verb conversions in contemporary Chinese, and categorizing these denominal usages under the same relation types described by \citet{clark1979nouns}. It was found that the eight major relation types of English denominal verbs can explain most of their Chinese counterparts, despite some small differences. We therefore extend our probabilistic framework of English noun-to-verb conversion to model how Chinese speakers might comprehend and produce denominal verb usages, hence testing the generality of our proposed framework to represent denominal meaning in two very different languages. 

Similar to DENOM-ENG, we performed an online corpus search on the iWeb-2015-Chinese corpus via Sketch Engine to determine the top 3 most common paraphrase verbs for each Chinese denominal utterance. This frequency-based corpus search yields a supervised set of 230 Chinese denominal utterances. We also augmented DENOM-CHN by replacing the denominal verb $D$ of each $U$ in \citet{Bai2014DenominalVI} with a set of synonyms taken from the taxonomy of Chinese Open WordNet database \cite{wang2013building}. After excluding cases with morphological or tonic changes during noun-to-verb conversions, we obtained an unsupervised set of 235 denominal utterances.

\subsection{Denominal verb usages in historical development of English (DENOM-HIST)}
To determine English nouns that had a temporal noun-to-verb conversion in history, we used the syntactically parsed Google Books Ngram Corpus that contains the frequency of short phrases of text (\textit{ngrams}) from books written over the past two centuries \cite{goldberg2013dataset}. 

We first extracted time series of yearly counts for words (\textit{1-grams}) whose numbers of occurrence as nouns and verbs both exceed a frequency threshold $\theta_f$, and we computed the proportion of noun counts for each word $w$ as follows:

\begin{equation}
    Q(w,t) = \frac{\#(w \text{ as a noun at year $t$})}{\#(w \text{ as a noun at year $t$}) + \#(w \text{ as a verb at year $t$})}
\end{equation} 

We then applied the change-point detection algorithm introduced by \citet{kulkarni2015statistically} to find words with a statistically significant shift in noun-to-verb part-of-speech (POS) tag ratio. This method works by detecting language change over a general stochastic drift and accounting for this by normalizing the POS time series. The method identifies change points via bootstrapping under a null hypothesis that in most cases, the expected value of a word's POS percentage should remain unchanged (compared to random fluctuations). Therefore, by  permuting the normalized POS time series, the pivot points with the highest shifts in mean percentage would be the statistically  significant change points.  Applying this method yielded a set of 57 target words as denominal verbs for our diachronic analysis. Since the n-gram phrases in Google Syntactic-Ngrams (GSN) are too short (with maximum length of 5 words) to extract complete denominal utterances and paraphrases, we considered another historical English corpus, the Corpus of Historical American English (COHA), that comprises annotated English sentences from 1810s to 2000s. We assumed that each denominal verb $w$ has been exclusively used as a noun prior to $t^*(w)$, and we extracted paraphrase usages $I^{(t)}$ before $t^*(w)$ as conventional usages, and denominal utterances $U^{(t)}$ after $t^*(w)$ as novel usages for prediction. All denominal utterances and paraphrases with aligned contextual objects and targets are taken as the supervised set, while the denominal utterances without aligned paraphrases found in the historical corpus are used for unsupervised learning, yielding an $X_s$ of size 1,055 and an $X_u$ of size 8,972.

\subsection{Crowd-sourced annotation of denominal verb usages  (DENOM-AMT)}
We evaluate our models on a set of denominal utterances with high-quality ground-truth paraphrases interpreted by human annotators. We collected human interpretations for a subset of the English and Chinese denominal verb usages in the training set described above via Amazon Mechanical Turks (AMT) crowdsourcing platform. 

For each utterance $D$, we presented the online participants with the top 3 paraphrase verbs collected from the iWeb corpora via frequency-based search, and asked the participants to choose, among the 3 candidates, all verbs that serve as good paraphrases for the target denominal verb in the denominal utterance. If none of them is appropriate, then the participants must provide a good alternative paraphrase verb by themselves. All annotators of English and Mandarin Chinese denominal verb examples must have passed a qualification test to confirm their proficiency in the respective languages to participate in the annotation.\footnote{See \url{https://github.com/jadeleiyu/noun2verb/tree/main/data/annotations} for questionnaires of language proficiency test and denominal utterance interpretation.} This online crowdsourcing procedure yields 744 annotated examples in English and 55 examples in Chinese (24 English utterances in DENOM-ENG were discarded due to insufficient number of collected responses).\footnote{We did not collect human responses for all examples in DENOM-CHN because many denominal uses have become obsolete in contemporary Mandarin (though they still appear in formal text such as official documents and therefore can be found via web corpus search). The first author therefore manually selected 54 Chinese denominal verbs considered to have nominal meanings familiar to modern Mandarin speakers.} For each utterance in English, there are on average 14.7 responses and 2.43 unique types of paraphrase verb collected, while for Chinese we got 12.8 responses and 1.97 paraphrase verb types per utterance. The resulting dataset includes in total 606 unique types of English denominal verbs and 54 unique types of Chinese denominal verbs. The English annotators reached an agreement score of $\kappa=0.672$ measured with Cohen's Kappa, and $\kappa=0.713$ for Chinese annotators. For English questions, 407 out of 744 denominal utterances have at least one alternative paraphrases provided in the annotations; for Chinese questions, 19 out of 55 utterances have at least one alternative paraphrase.


\section{Evaluation and results}


We first describe the experimental details and the procedures for evaluation of our proposed framework. We then present three case studies that evaluate this framework against different sources of innovative denominal usages drawn from speakers of different age groups and across languages, as well as data from contemporary and historical periods.

\subsection{Details of experimentation and evaluation}

We ran the proposed probabilistic models on the 3 training datasets (DENOM-ENG, DENOM-CHN and DENOM-HIST) by optimizing over their loss functions specified in Section 2. The speaker and listener modules in partial and full generative models are implemented as three-layer feed-forward neural networks using the Pyro deep probabilistic programming library \cite{bingham2018pyro}. For the discriminative model in the contemporary datasets, we initialized both the listener and speaker modules with 12-layer pre-trained BERT neural language models implemented by the HuggingFace library based on PyTorch, and we fine-tuned the parameters in BERT during training. The input sequences ($I,E$ for listener modules, and $U$ for speaker modules) were first encoded via distributed word embeddings, which were then fed into the corresponding modules for further computation. For synchronic prediction, we applied the GloVe algorithm \cite{pennington2014glove} on Wikipedia 2014 and Gigaword 5 corpora to learn distributed word embeddings. 

To initialize the models and prevent these embeddings from smuggling in information about target denominal verb usages for model prediction, we removed all denominal usages for target denominal verbs during training. For historical prediction, we replaced GloVe embeddings with the HistWords historical word embeddings used in \citet{hamilton2016diachronic} for each decade from 1800s to 1990s. Similar to the synchronic case, we re-trained all historical embeddings by explicitly removing all denominal usages of each target word $D$ (that we seek to predict) from the original text corpora.\footnote{We validated the reliability of the POS tagger by asking human annotators on AMT to manually inspect 100 randomly-sampled denominal utterances detected by SpaCy from the iWeb-2015 corpus. We collected 5 responses for each utterance, and found that at least 3 annotators agree with the automatically labeled POS tags for 94 out of 100 cases.}

We assess each model on the evaluation set of denominal verb usages that have ground-truth paraphrases, in the two types of predictive tasks described. In the comprehension tasks, for each novel denominal utterance $U$, we sample interpretation from the listener module's posterior distribution $p_l(I|U)$, and compare these model predictions against the ground-truth paraphrases provided by human annotators. In the production tasks, we conversely group all denominal utterances that share the same verb-relation pair as ground-truth interpretations of the intended meaning (e.g., ``mail my resume'' and ``email my number'' with common paraphrase verb ``send'' and relation ``INSTRUMENT''). For every interpretation, we apply the speaker module to generate novel denominal usages from the posterior distribution $p_s(U|I)$.

We consider two metrics to evaluate model performance: 1) Standard receiver operating curves (ROCs), which provide a comprehensive evaluation for the model prediction accuracy based on $k=1,2,3,...$ guesses of interpretation/utterances from its posteriors. Prediction accuracy (or precision) is the proportion of interpretations/utterances produced by the model that fall into the set of ground-truths---this metric automatically accounts for model complexity and penalizes any model that has poor generalization or predictive ability;  we also report the mean accuracy when considering only the top-$k$ model predictions from $k=1$ up to $k=5$. 2) Kullback–Leibler divergence $D_\text{KL}$, on the other hand, measures the ability of the models to capture fine-grained human annotations. Since each query denominal verb usage has multiple ground-truths (i.e., the set of paraphrases provided by human annotators that forms a ground-truth distribution), we compute the discrepancy between the empirical distributions $p_\text{emp}(U|I)$, $p_\text{emp}(I|U)$ of paraphrases/utterances collected from AMT workers, and model-predicted posteriors $p_l(I|U)$ and $p_s(U|I)$---a smaller $D_\text{KL}$ indicates better alignment between the natural distribution of human judgement and model posterior distribution (see Appendix D for details of calculating the KL divergence). Since the value of KL divergence may be sensitive to the size of the evaluation set, we calculate the $D_{KL}$ for contemporary English examples by randomly sampling from DENOM-AMT 100 subsets of English denominal utterances with the same size of the Chinese evaluation set, and taking the mean KL divergence between the sampled sets of utterances and their ground-truth paraphrases.

For the case of contemporary English, since almost all supervised examples are human-annotated, we adopt a 12-fold leave-one-out cross-validation procedure: first, we randomly split the 744 annotated utterances into 12 equally-sized subsets; second, we draw 11 subsets of annotated examples for model training (together with unsupervised utterances for generative models), and use the left out subset for model evaluation. We repeat this procedure 12 times and report the mean model performance. For the case of contemporary Chinese, we train models using all denominal utterances in DENOM-CHN (i.e., 235 unannotated utterances and 230 annotated examples with paraphrases determined via corpus search), and evaluate the models using the 55 human-annotated Chinese denominal utterances in DENOM-AMT. For the diachronic analysis in English, we keep a subset of the supervised learning example $X_s$ as test cases (and remove them from training sets). We also consider two additional baseline models for evaluating if any of the three proposed probabilistic models can learn above chance, at all: 1) a frequency-based model that chooses the interpretation and denominal usage with highest empirical probability $p_\text{emp}(U|I)$ or  $p_\text{emp}(I|U)$ in the training dataset, and 2) a random-guess model that samples each linguistic component from a uniform distribution.

\subsection{Case study 1: Contemporary English}

\begin{figure}[h]
\includegraphics[width=1\linewidth]{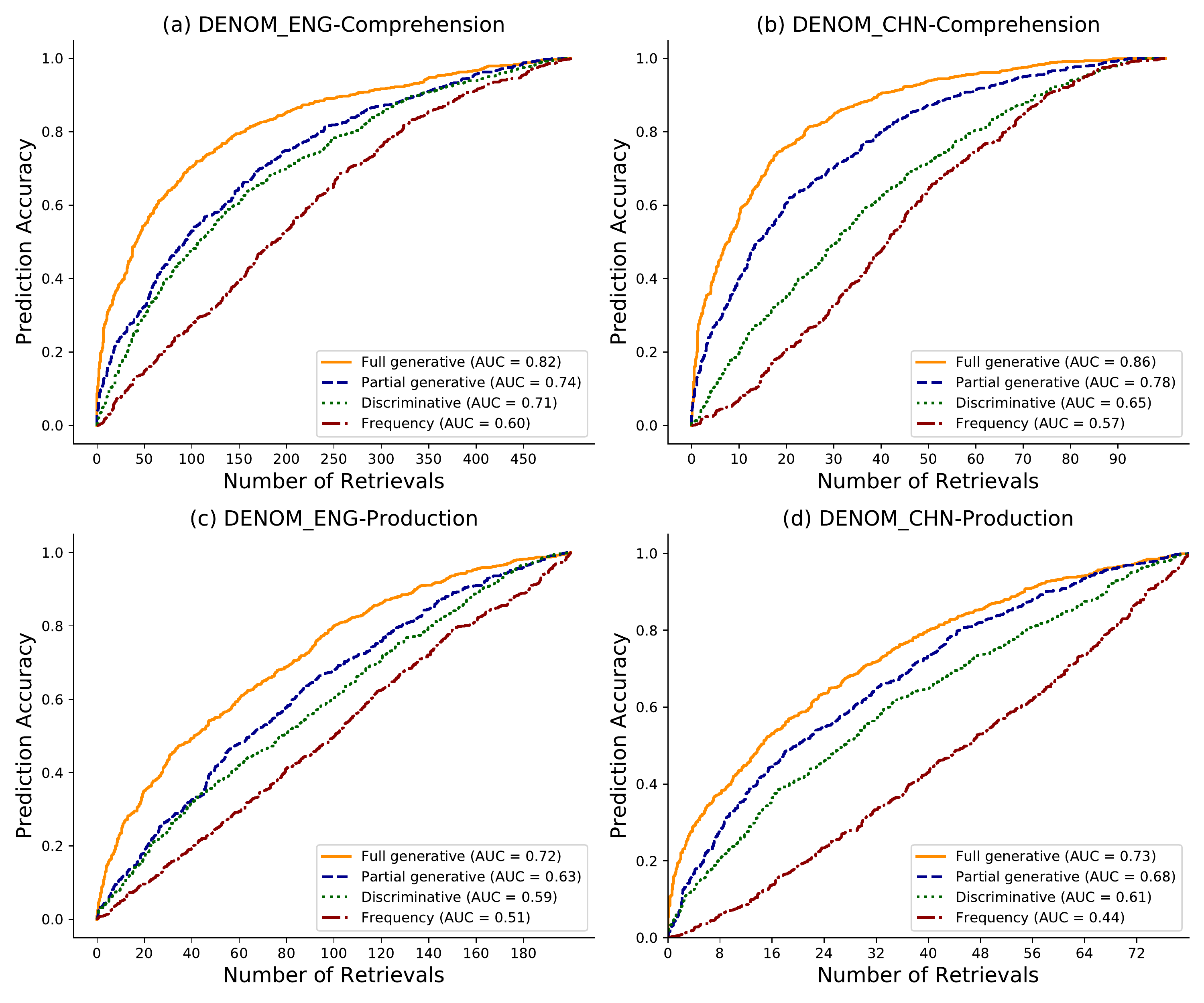}
\caption{A summary of model performance in denominal verb comprehension and production. The left column summarizes the results from the 744 English examples in the DENOM-AMT dataset based on receiver operating characteristic (ROC) curves, and the right column summarizes similar results from the 55 Chinese examples in DENOM-AMT dataset. ``Frequency'' refers to the frequency baseline model. Higher area-under-the-curve (AUC) score indicates better performance.}
\label{roc_curves}
\end{figure}

\begin{figure}[]
\includegraphics[width=1\linewidth]{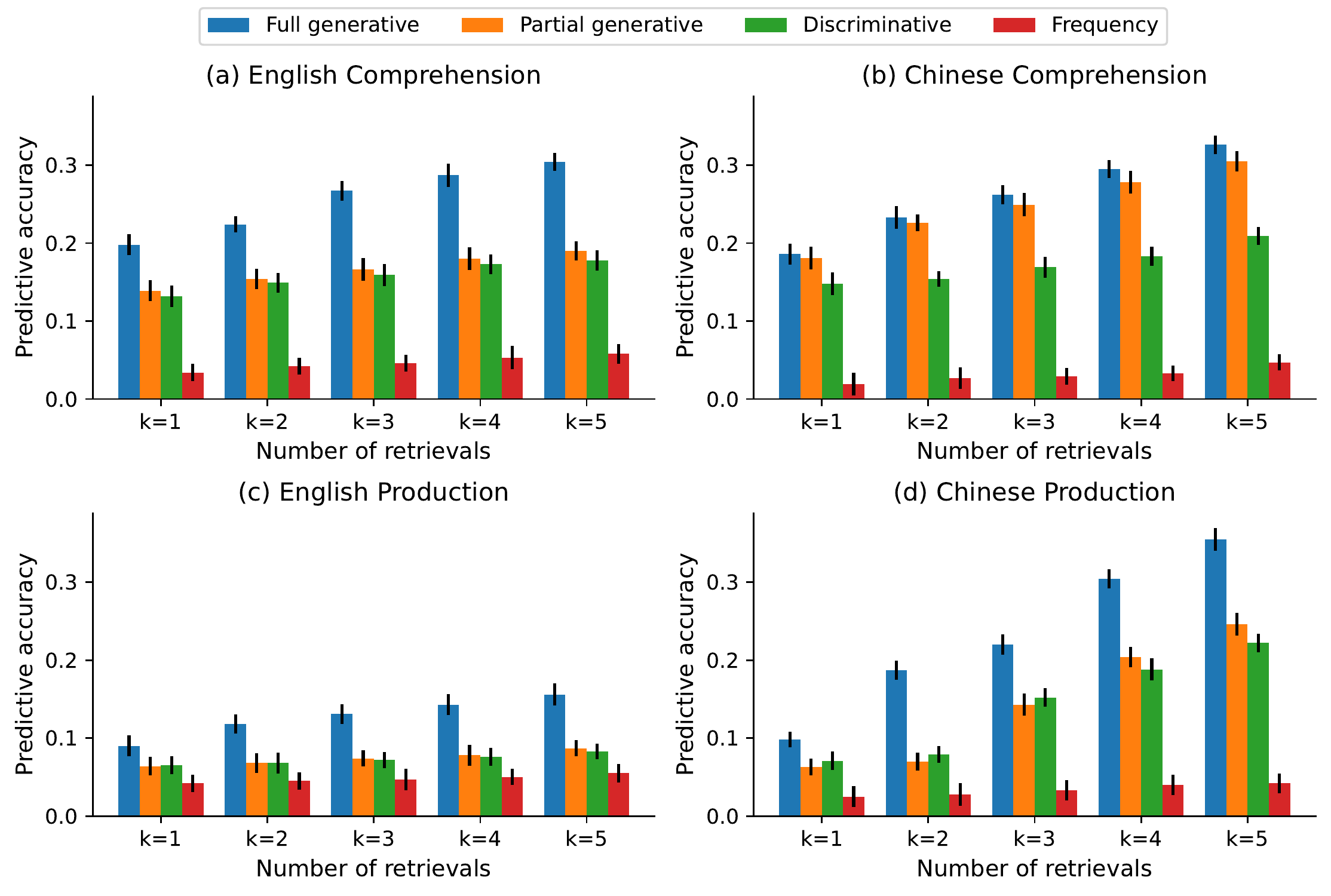}
\caption{Model predictive accuracy in denominal verb comprehension and production when taking the top-$k$ outputs, with $k$ ranging from 1 to 5. The left column summarizes the results from the 744 English examples in the DENOM-AMT dataset, and the right column summarizes similar results from the 55 Chinese examples in DENOM-AMT. ``Frequency'' refers to the frequency baseline model. Vertical bars represent standard errors.}
\label{topk-accs}
\end{figure}

We first evaluate the models in comprehension and production of contemporary English denominal verb usages. Figures~\ref{roc_curves}a and \ref{roc_curves}c summarize the results on 744 English denominal utterances in DENOM-AMT using ROCs, while Figures~\ref{topk-accs}a and \ref{topk-accs}c show the predictive accuracy on the same evaluation dataset when considering the top-$k$ outputs for each model, with $k$ ranging from 1 to 5. 

We  found  that  all  non-baseline  models  achieved good accuracy in predicting semantic relation types (lowest accuracy $96\%$), so we focused our discussion on model predictions of interpreting via paraphrased verbs $V$ and generating novel denominal verbs $D$. We  computed  the  area-under-the-curve  (AUC)  statistics to compare the cumulative predictive accuracy of the models, summarized also in Figure~\ref{roc_curves}.  Our full generative model yields the best AUC scores and top-$k$ predictive accuracy for both tasks, outperforming the partial generative model, which is in turn superior to the BERT-based discriminative model.

The left two columns of Table~\ref{table:kl-divs} show the mean KL-divergence scores between model posteriors and empirical distributions over both interpretations and denominal utterances on all 744 English test cases in DENOM-AMT. We observed that both full and partial generative models offer better flexibility in interpreting and generating novel denominal verb usages, but the discriminative model, despite its high predictive accuracy, yields output distributions that are least similar to human word choices among non-baseline models. In particular, we found that the generative models outperform their discriminative counterparts on both child and adult denominal utterances (see Appendix C for a detailed breakdown of these results).

\begin{table}[h]
\centering
\caption{Model comparison on predicting human annotated denominal data. Model accuracy is summarized by Kullback-Leibler (KL) divergence between posterior distributions $p_\text{comp}(V|U)$, $p_\text{prod}(D|I)$ and fine-grained empirical distributions of human-annotated ground-truth on DENOM-AMT dataset. A lower value in KL indicates better alignment between model distribution and empirical distribution. Standard errors are shown within the parentheses.}

\begin{tabular}{@{}lllll@{}}
\toprule
\multicolumn{1}{c}{}      & \multicolumn{4}{c}{KL divergence ($\times 10^{-3}$)}                               \\
\multicolumn{1}{c}{Model} & \multicolumn{2}{c}{English} & \multicolumn{2}{c}{Chinese} \\
                          & Comprehension   & Production   & Comprehension   & Production   \\ \midrule
Full Generative           & 8.86 (1.1)      & 21.7 (2.4)   & 2.93 (0.46)     & 7.8 (1.1)    \\
Partial Generative        & 10.01 (0.9)      & 22.4 (2.5)   & 3.08 (0.35)     & 11.0 (1.1)   \\
Discriminative            & 13.75 (1.0)      & 39.0 (1.8)  & 3.32 (0.33)     & 29.4 (1.8)   \\
Frequency Baseline        & 11.41 (0)        & 57.7 (0)    & 3.62 (0)        & 28.5 (0)     \\ \bottomrule
\end{tabular}

\label{table:kl-divs}
\end{table}

\begin{table}[ht]
\centering
\caption{Model comparison on predicting human-generated and corpus-generated denominal utterance paraphrases.}
\resizebox{0.75\textwidth}{!}{
\begin{tabular}{@{}lllll@{}}
\toprule
\multicolumn{1}{c}{}      & \multicolumn{4}{c}{Comprehension accuracy}                               \\
\multicolumn{1}{c}{Model} & \multicolumn{2}{c}{English} & \multicolumn{2}{c}{Chinese} \\
                          & Human   & Corpus   & Human   & Corpus   \\ \midrule
Full Generative           & 0.406      & 0.459   & 0.279     & 0.288    \\
Partial Generative        & 0.373      & 0.408   & 0.243     & 0.257   \\
Discriminative            & 0.297      & 0.462  & 0.222     & 0.276   \\
Frequency Baseline        & 0.153        & 0.307     & 0.097       & 0.109     \\ \bottomrule
\end{tabular}
}
\label{table:human-vs-corpus-paraphrase-accs}
\end{table}

\begin{CJK*}{UTF8}{gbsn}
\begin{figure}[h]
\includegraphics[width=1\linewidth]{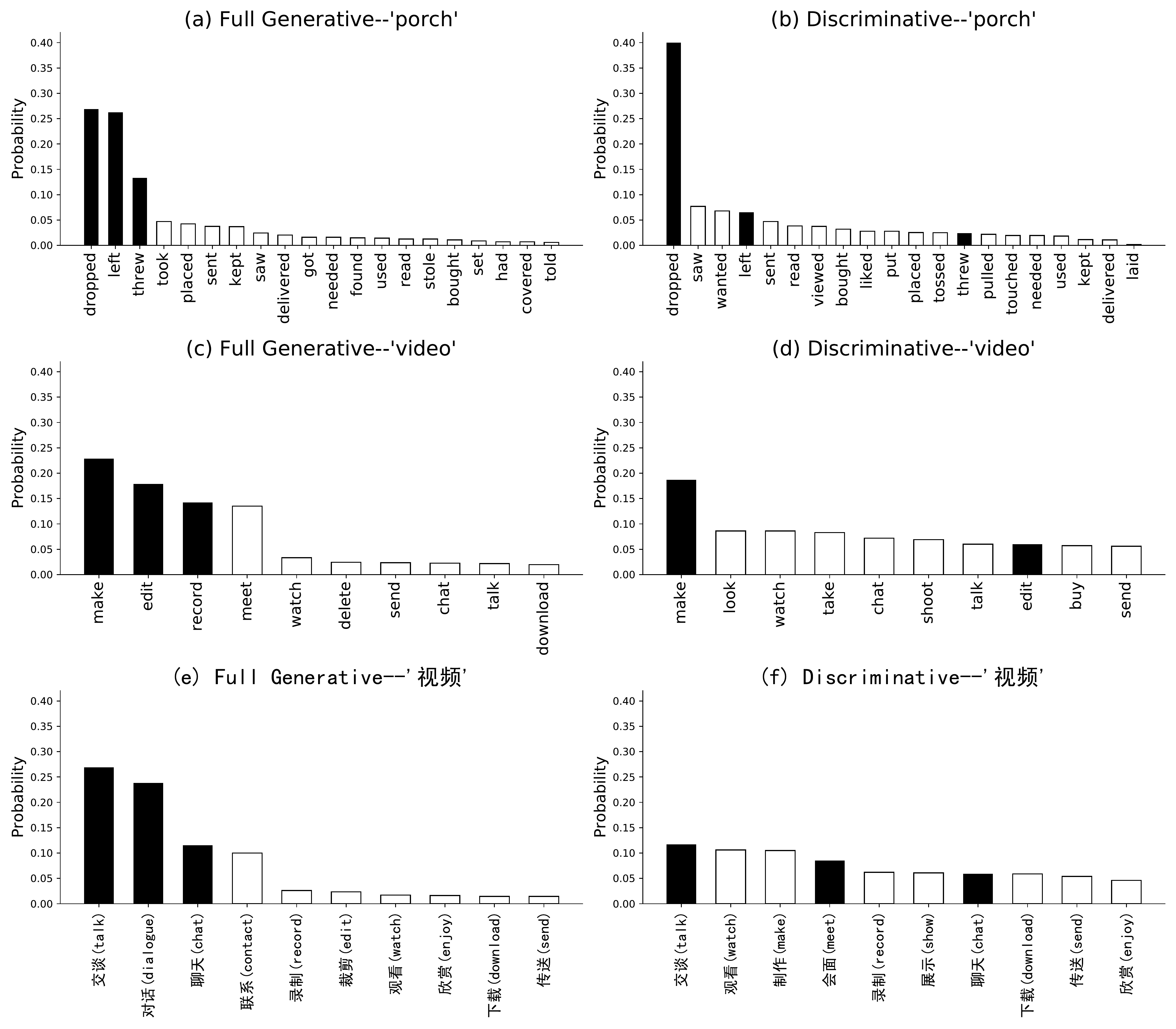}
\caption{Comparison of the full generative model and the discriminative model on the quality of paraphrasing novel denominal usages. The top 2 rows show model posterior distributions on the paraphrase verbs (horizontal axis) for the query denominal verbs {\it porch} and {\it video} in English. The bottom row shows similar information for the query denominal verb {\it video-chat} in Chinese, with the English translations in the parentheses. Model predicted probabilities for the top 3 choices from human annotation (i.e., ground truth) are shown in solid black bars.}
\label{posterior_examples}
\end{figure}
\end{CJK*}

To demonstrate the better flexibility of the full generative model, we visualize in the first row of Figure~\ref{posterior_examples} the listener posterior distributions $p_l(V|U)$ over paraphrase verbs for the query denominal usage $U = $``porch the newspaper'' based on the full generative and discriminative models (top 20 candidates with non-zero probabilities are shown). We found that the full generative model assigned the highest posterior  probabilities  on  the  three  ground-truth human-annotated verb paraphrases \textit{dropped}, \textit{left} and \textit{threw}, and the partial generative model also ranked them as the top 5 candidates (posterior of which is not shown in the figure). In contrast, the discriminative model only assigned the highest posterior probability for \textit{drop}, and failed to distinguish the two alternative ground-truths between other implausible candidate paraphrase verbs. The second and third most likely candidates predicted by the discriminative model are \textit{saw} and \textit{wanted}, most possibly because these are commonly associated words in the pre-training text corpora of BERT. This limitation of not being able to explain the fine-grained distribution of paraphrases and only locking onto a single best solution has appeared to be a general issue for the discriminative model, as we observed the same phenomenon in many other test cases.

We also examine whether our generative models outperform the baselines in the comprehension tasks by simply favoring paraphrase verbs that are more frequently paired with the target denominal verbs in the linguistic corpora. The left two columns of Table ~\ref{table:human-vs-corpus-paraphrase-accs} summarize individual model predictive accuracy on interpreting denominal utterances in DENOM-AMT which have ground-truth paraphrases  either completely generated by corpus search (i.e., no alternative paraphrases collected) or produced as alternative paraphrases by human annotators. For corpus-generated cases, we calculate the model comprehension accuracy when considering the top-3 output paraphrase verbs; for human-generated cases with $m$ alternative interpretations, we calculate the model comprehension accuracy when considering the top-$m$ output paraphrase verbs. We found that the full generative model offers most accurate interpretations in both cases, while the discriminative model, despite its high accuracy in predicting corpus-generated paraphrases, performs significantly worse on human-generated examples. These results further demonstrated the inflexibility of discriminative language models on interpreting denominal utterances, especially when the feasible paraphrase verbs rarely co-occur with a given target noun in the reference corpora. 

These initial findings on contemporary English data therefore suggest a generative, frame-based approach to denominal verbs that encodes speaker-listener shared knowledge and latent semantic frames.

\subsection{Case study 2: Contemporary Chinese}

\begin{CJK*}{UTF8}{gbsn}
We next investigate whether the same framework generalizes to denominal verb usages in a language that is markedly different from English. Table~\ref{chinese_examples} presents some exemplar denominal verbs that are common in contemporary Mandarin Chinese. Although Chinese does not have morphological markings for word classes, there exist alternative linguistic features that signify the presence of a verb. For example, a word that appears after the auxiliary word ``地'' or before ``得'' is typically a verb. If a word that is commonly used as a noun appears in context of these verbal features, it can then be considered as a denominal verb. For example, the phrase ``开心地\_\_'' denotes ``to \_\_ happily'', where ``\_\_'' is filled with a verb. Therefore, when a  noun such as ``视频'' (video) appears in the phrase ``开心地视频'' (to video happily), a Chinese speaker would consider ``视频'' as a verb converted from its nominal class. It is worth noting that for some Chinese nouns, their direct translations into English are still valid denominal verbs, but their meaning may differ substantially in two languages. For instance, the denominal utterance ``I videoed with my friends'' would remind an English speaker of a scenario of video recording, while for Chinese speakers the correct interpretation should be ``I chat with my friends via online video applications''. We therefore expect our models to be able to capture such nuanced semantic variation when learning from the Chinese denominal dataset DENOM-CHN. 
\end{CJK*}

The right columns of Figure~\ref{roc_curves} and Figure ~\ref{topk-accs} show respectively the ROC curves (with AUC scores) and the top-$k$ predictive accuracy of each model on the comprehension and production tasks in Mandarin Chinese. We observed that similar to the case of English denominal verbs, the full generative model yields the best overall predictive accuracy. Moreover, as shown in the right two columns of Table~\ref{table:kl-divs}, the generative model aligns better with Chinese speakers' interpretation and production of denominal verbs, because it yields a lower KL-Divergence score between its posteriors and empirical distributions of ground-truths in comparison to the discriminative model. Moreover, as shown in the right two columns of Table~\ref{table:human-vs-corpus-paraphrase-accs}, the performance of the discriminative still drops significantly when switching from predicting corpus-generated to human-generated paraphrases, while its generative counterparts are much less influenced by this change. 

\begin{CJK*}{UTF8}{gbsn}
As an example where our framework successfully captures cross-linguistic semantic variation in denominal verbs, the second and third rows of Figure~\ref{posterior_examples} illustrate the posterior distributions over paraphrase verbs for the discriminative and full generative models on the exemplar utterance ``to video (视频) with my friends'', in English and Chinese. In both languages, the full generative model assigns the highest probability masses on the three ground-truth paraphrases chosen by human annotators, thus demonstrating its ability to flexibly interpret a denominal verb based on its linguistic context under different languages. The discriminative model not only favors idiosyncratically a single ground-truth verb (``
交谈(talk)'' for Chinese and ``make'' for English), but it also yields less flexible model posteriors when translating from English to Chinese. For instance, the BERT model fails to realize that ``make'' should no longer be a good paraphrase in Chinese, while the full generative model successfully excludes this incorrect interpretation from the top 10 candidates in the listener's posterior distribution. 
\end{CJK*}

\begin{CJK*}{UTF8}{gbsn}
\begin{table}[h]
\caption{Examples of denominal verb usages in contemporary Mandarin Chinese, together with their literal translations and interpretations in English. The target Chinese denominal verbs and their English translations are underlined, and their corresponding English paraphrase verbs are marked in bold font.}
\resizebox{\textwidth}{!}{%
\begin{tabular}{@{}cll@{}}
\toprule
Chinese denominal verb usage & Literal translation in English & Interpretation in English       \\ \midrule
\underline{漆}门窗                          & \underline{paint} doors windows            & to \textbf{paint} doors and windows      \\
\underline{网}鱼                           & \underline{net} fish                       & to \textbf{catch} fish with the net      \\
\underline{圈}地                           & \underline{circle} land                    & to \textbf{enclose} land                 \\
和朋友\underline{视频}                        & with friends \underline{video}             & to \textbf{chat} with friends via webcam \\ \bottomrule
\end{tabular}%
}
\label{chinese_examples}
\end{table}
\end{CJK*}

These results further support our generative, frame-based approach to denominal verbs that explains empirical data in two different languages.

\subsection{Case study 3: Historical English}

In our final case study, we examine whether our framework can predict the emergence of novel denominal verb usages in the history of English.

\begin{figure}[t]
\includegraphics[width=1\linewidth]{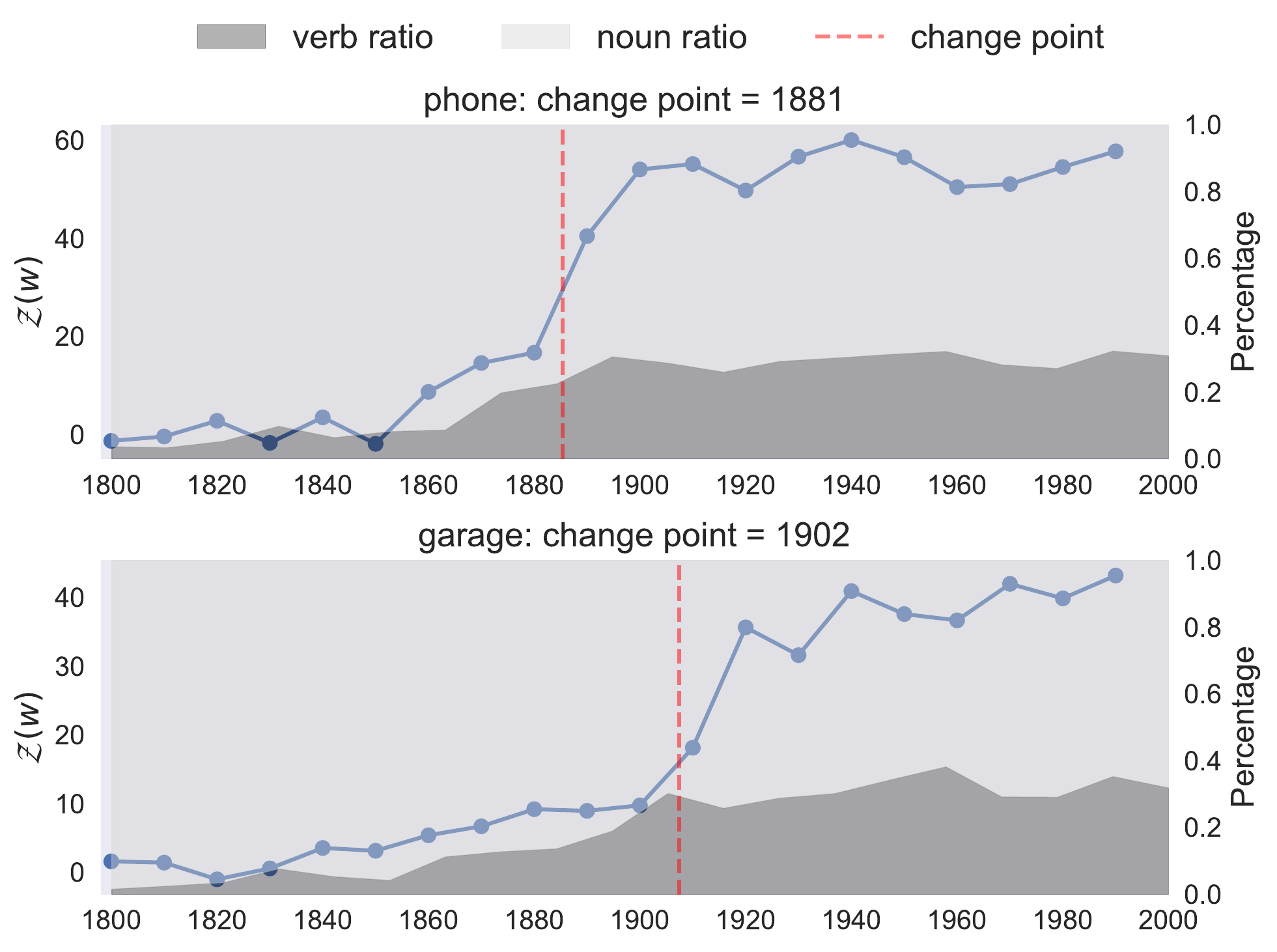}
\caption{Frequency-based z-score time series $Z(w) = Z(Q(w,t))$ for the nouns ``phone'' and ``garage'' over the past two centuries. The stacked color areas denote percentage of noun/verb usage frequencies in each year, and the red vertical lines mark the detected historical change point of noun-to-verb conversion.}
\label{change_points}
\end{figure}

We first demonstrate the effectiveness of our change point detection algorithm in determining valid noun-to-verb conversions in history. Figure~\ref{change_points} shows the word frequency-based Z-score series $Z(w)$ for the noun-to-verb count ratio series $Q(w,t)$ of some sample words, together with the change points $t^*(w)$ detected by our algorithm. We observed that the algorithm correctly identifies substantial shifts in noun-to-verb count ratios across time.

\begin{figure}[t]
\includegraphics[width=1\linewidth]{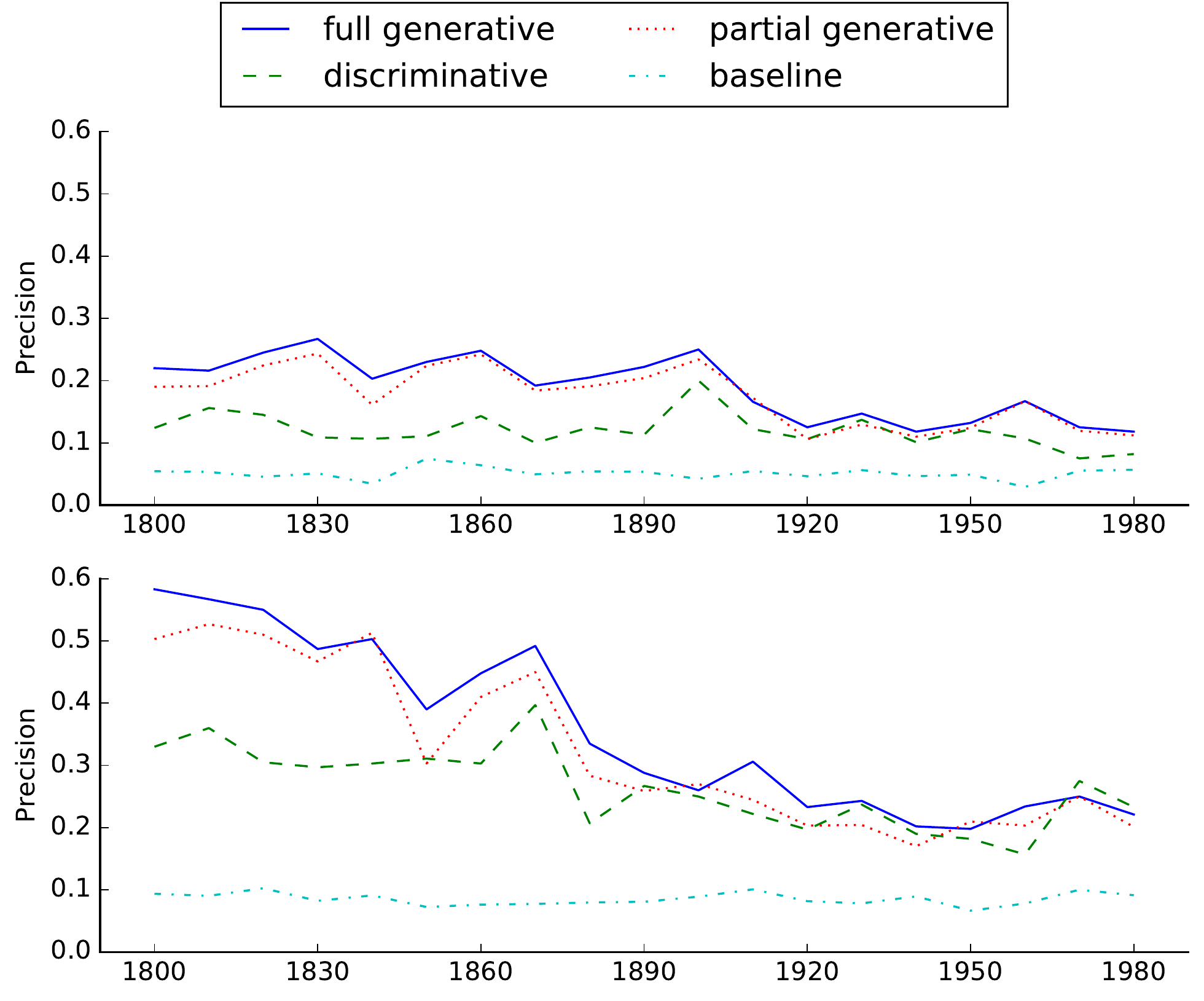}
\caption{Model predictive accuracy on DENOM-HIST dataset. Top row shows the predictive accuracy where only emerging denominal verb usages in the immediate next decade are considered. Bottom row shows predictive accuracy where all future emerging denominal usages are considered.}
\label{temp_accs}
\end{figure}

We next report the temporal predictive accuracy of our models by dividing the evaluation set of denominal verbs into groups where change points fall under the same decade. For each target word $D$ with $m$ novel denominal usages observed after its detected change point $t^*(w)$, we take the top $m$ sampled usages with the highest speaker's posterior probability $p_{s}(U|I)$ as the set of retrieved model predictions, and we then calculate the average predictive accuracy of the top $m$ predictions over all denominal verbs emerged in each decade. We considered two kinds of evaluation criteria when making predictions for target $D$ in future decade $t$: taking as ground truth denominal utterances that contain $D$ (1) specifically in the immediate following decade $t + 1$, and (2) any future decade $t' > t$ up to the terminal decade 2000s.

Figure~\ref{temp_accs} shows the decade-by-decade precision accuracy for all the three probabilistic models, along with a frequency baseline that always chooses the top $m$ denominal utterances with the highest frequencies that contain $D$. The predictive accuracy falls systematically in later decades because there are fewer novel denominal verbs to predict in the data. To ensure that most target nouns have attested denominal uses in future decades, we only calculate model precision up to 1980s.   The full generative model yields consistently better results in almost every decade, and it is followed by the BERT-based discriminative model. Note that in later decades, the difference in accuracy of BERT model and the full model becomes smaller, presumably due to the increasing similarity between learning data and text corpora on which BERT is pre-trained (i.e., contemporary text corpora). Overall, our generative model yields  more accurate prediction on denominal verb usages than the discriminative model with rich linguistic knowledge. 

Taken together, these results provide firm evidence that our probabilistic framework has the explanatory power over the historical emergence of denominal verb usages. 

\section{Model interpretation and discussion}

To interpret the full generative model, we visualize the latent frame representations learned by this model. We also discuss the strengths and limitations from qualitative analyses of example denominal verb usages interpreted and generated by the model.

\begin{figure}[ht]
\includegraphics[width=1\linewidth]{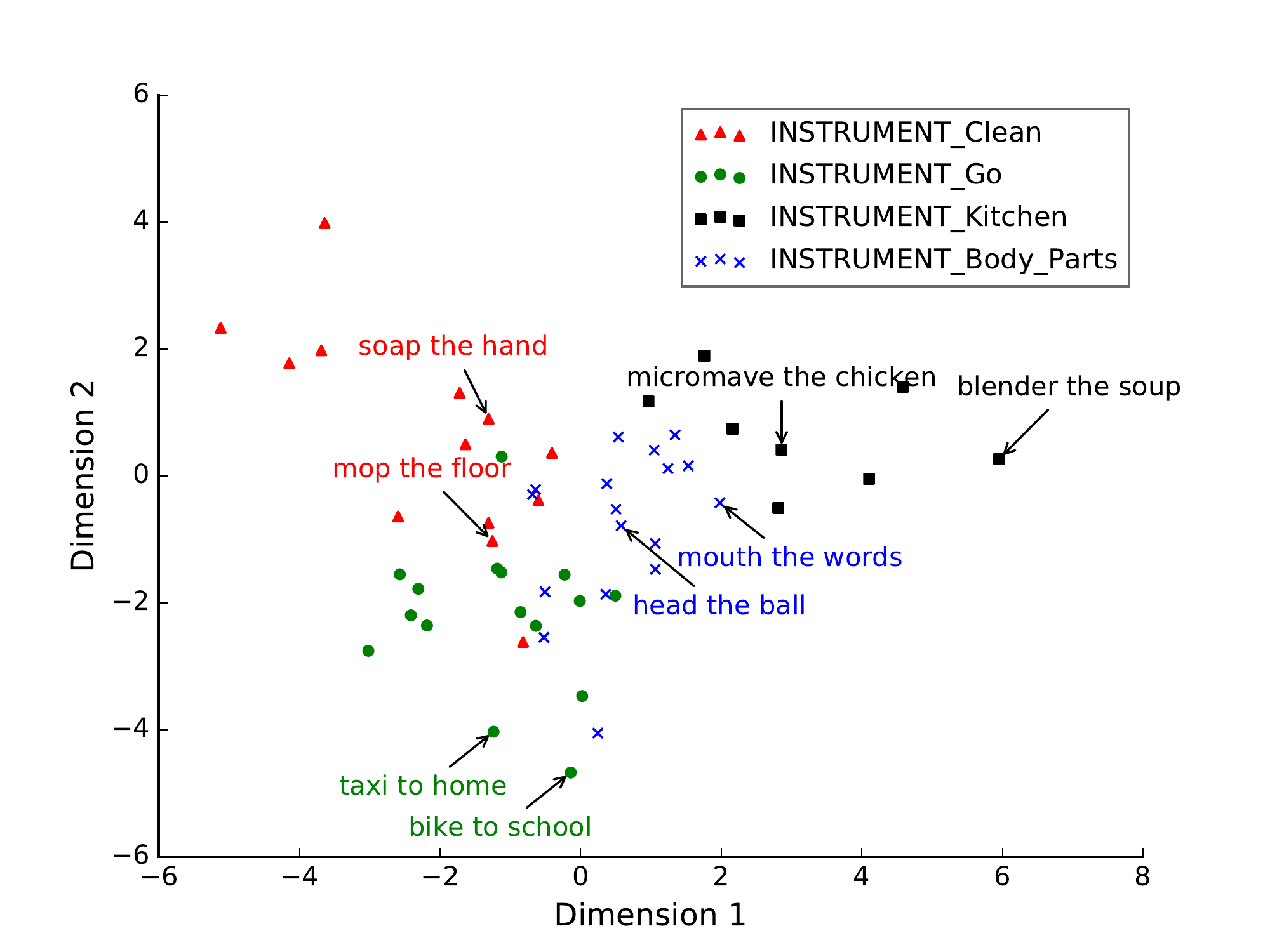}
\caption{t-distributed Stochastic Neighbor Embedding visualization of the model-learned frame elements ($E$) for denominal verb usages in the category of INSTRUMENT (DENOM-ENG dataset). Utterances within each sub-category based on~\cite{clark1979nouns} are shown with the same marker. For each sub-category, the most frequent denominal utterances are annotated with its source text.}
\label{frame_tsne}
\end{figure}

Classic work by \citet{clark1979nouns}  provided a fine-grained classification of denominal usages within the relation type INSTRUMENT. We use this information to gain an intuitive understanding on the full generative model in its ability to capture fine-grained semantics via the frame element variable $E$. Figure~\ref{frame_tsne} shows the t-distributed Stochastic Neighbor Embeddings (t-SNE) \cite{maaten2008visualizing}, a nonlinear dimensionality reduction method that projects high-dimensional data into low-dimensional space for visualization, of the model-learned latent frame elements for the example denominal utterances in INSTRUMENT type. All data points are expressed in  markers following their sub-category labels pre-specified in \cite{clark1979nouns}. We observe that the learned frame variable $E$ encourages denominal usages within the same sub-category to be close in semantic space, manifested in the four clusters. As such, the frame variable helps to capture the fine-grained distinctions of denominal usages (even within the same broad semantic category).

\begin{CJK*}{UTF8}{gbsn}
\begin{table}[]
\caption{Example paraphrases of novel denominal usages interpreted by the full generative model.}
\resizebox{\textwidth}{!}{%
\begin{tabular}{lllll}
\hline
\textbf{\begin{tabular}[c]{@{}l@{}}Denominal\\ usage\end{tabular}} & \textbf{Dataset} & \textbf{\begin{tabular}[c]{@{}l@{}}Semantic \\ relation\end{tabular}} & \textbf{\begin{tabular}[c]{@{}l@{}}Human paraphrases \\ with model-predicted ranks in ()\end{tabular}} & \textbf{\begin{tabular}[c]{@{}l@{}}Top verb paraphrases \\ inferred from the model\end{tabular}} \\ \hline
\underline{blanket} the bed                                                       & DENOM-ENG        & locatum on                                                            & put(1), place(3), lay(11)                                                                              & put, drop, cover                                                                                 \\
\underline{paper} my hands                                                         & DENOM-ENG        & instrument                                                            & cut(1), hurt(2)                                                                                        & cut, hurt, wound                                                                                 \\
\underline{fox} the police                                                         & DENOM-ENG        & agent                                                                 & deceive(4), baffle(2), fool(3)                                                                         & cheat, baffle, fool                                                                              \\
\underline{网}鱼 (\underline{net} the fish)                                                      & DENOM-CHN        & instrument                                                            & 捕(catch, 1), 抓(capture, 3)                                                                             & 捕(catch), 捉(seize), 抓(capture)                                                                   \\
\underline{garage} the car                                                         & DENOM-HIST       & location on                                                           & stop(1), put(6), park(2)                                                                               & stop, park, move                                                                                 \\ \hline
\underline{mine} the gold                                                          & DENOM-ENG        & location out                                                          & dig(327), extract(609), get(25)                                                                        & put, bury, find                                                                                  \\
\underline{bee} the cereal                                                         & DENOM-ENG        & locatum on                                                            & add(54)                                                                                                & get, find, eat                                                                                  
\end{tabular}%
}
\label{samples-comprehension}

\caption{Examples of novel denominal usages produced by the full generative model.}
\resizebox{\textwidth}{!}{%
\begin{tabular}{lllll}
\hline
\textbf{\begin{tabular}[c]{@{}l@{}}Paraphrase \\ verb\end{tabular}} & \textbf{Dataset} & \textbf{\begin{tabular}[c]{@{}l@{}}Semantic \\ relation\end{tabular}} & \textbf{\begin{tabular}[c]{@{}l@{}}Ground-truth \\ denominal verb utterances\end{tabular}}       & \textbf{\begin{tabular}[c]{@{}l@{}} Denominal usages sampled\\ from model posterior\end{tabular}} \\ \hline
remove                                                                 & DENOM-ENG        & instrument                                                            & \begin{tabular}[c]{@{}l@{}}shell the peanuts, fin the fish, \\ skin the rabbit\end{tabular} & stem the flowers                                                                                     \\
repeat                                                              & DENOM-ENG        & agent                                                                 & parrot my words                                                                             & chimpanzee my gestures                                                                              \\
选(choose)                                                           & DENOM-CHN        & instrument                                                            & 筛人(sieve the candidates)                                                                    & 筛选书籍(sieve the books)                                                                                \\
冷却(freeze)                                                          & DENOM-CHN        & instrument                                                            & 冰水(ice the water)                                                                           & 冰食物(ice the food)                                                                                   
\end{tabular}%
}
\label{samples-production}
\end{table}
\end{CJK*}

To gain further insight into the generative model, we show in Tables~\ref{samples-comprehension} and \ref{samples-production}  example model predictions in the three datasets we analyzed, for denominal verb comprehension and production respectively. In the comprehension task (Table~\ref{samples-comprehension}), our model provides reasonable interpretations on novel denominal verbs that did not appear in the learning phase. For instance, the model inferred that {\it blanket the bed} can be paraphrased as ``put/drop/cover the blanket on the bed'', which are close approximations to the top 3 ground-truth paraphrases ``put/place/lay the blanket on the bed''. One factor that facilitated such inference is that there are analogous denominal verbs during model learning, e.g., {\it carpet the floor}, that allowed the model to extrapolate to new denominal cases.

In the production task (Table~\ref{samples-production}), our model successfully generated both 1) now conventionalized denominal usages such as {\it stem the flowers}, even though {\it stem} was completely missing in the learning data, and 2) truly novel denominal cases such as {\it chimpanzee my gestures}, presumably by generalization from training examples such as {\it parrot my words}.

Similar to the case of English, models trained on Chinese denominal data also exhibit generalizability. However, we also observed poor model generalization, especially when training instances from a semantic type are extremely sparse. For instance, as \citet{clark1979nouns} point out, very few English denominal verbs fall under the ``location out'' relation type, and our model therefore often misinterpreted denominal usages under this type (e.g., {\it mine the gold}) as cases from ``location in'' type (e.g., ``put/bury/find the gold in(to) the mine''). These failed cases suggest that richer and perhaps more explicit ontological knowledge encoded in denominal meaning, such as the fact that mines are places for excavating minerals, should be incorporated.
These failed cases might also be attributed to the difficulty in distinguishing different types of semantic relations with word embeddings, such as synonyms from antonyms.

Another issue concerns the semantic representation of more complex denominal verbs that cannot be simply construed via paraphrasing, but are otherwise comprehensible for humans. For example, the denominal usage {\it bee the cereal} was observed in a sub-corpus of the CHILDES dataset, where a child asked the mother to add honey to his cereal. Interpreting such an innovative utterance requires complex (e.g., a chain of) reasoning by first associating bees with honey, and then further linking honey and cereal. Since the paraphrase templates we worked with do not allow explanations such as ``to add the honey (produced by bees) into the cereal'', all models failed to provide reasonable interpretations for this novel usage. 

Our framework is not designed to block novel denominal usages. Previous studies suggest that certain denominal verbs are not productive due to blocking or statistical preemption, e.g., we rarely say {\it car to work} because the denominal use of {\it car} is presumably blocked by the established verb {\it drive}~\cite{clark1979nouns,goldberg2011corpus}. We believe that this ability of knowing what not to say cannot be acquired without extensive knowledge of the linguistic conventions of a language community, although we do not consider this aspect as an apparent weakness of our modeling framework (since {\it car to work} does have sensible interpretations for English speakers, even though it is not a conventional expression in English). In contrast, we think it is desirable for our framework to be able to interpret and generate such pre-emptive cases, because such expressions though blocked in one language can be productive and comprehensible in other languages. Our generative models are able to produce and interpret such potential overgeneralizations due to their exposure to unsupervised denominal utterances generated from synonym substitutions as explained in Section 4.1.

\section{Conclusion}
We have presented a formal computational framework for word class conversion with a focus on denominal verbs. We formulate noun-to-verb conversion as a dual comprehension and production problem between a listener and a speaker, with shared knowledge represented in latent semantic frames. We show in an incremental set of probabilistic models that a generative model encoding a full distribution over semantic frames best explained denominal usages in contemporary English (by adults and children), Mandarin Chinese, and the historical development of English. Our results confirmed the premise that probabilistic generative models, when combined with structured frame semantic knowledge, can capture the comprehension and production of denominal verb usages better than discriminative language models. Future work can explore the generality of our approach toward characterizing other forms of word class conversion. Our current study lays the foundation for developing natural language processing systems toward human-like lexical creativity.

\begin{acknowledgments}
We thank Graeme Hirst and Peter Turney for helpful feedback on our
manuscript. We thank Lana El Sanyoura, Bai Li, and Suzanne Stevenson for
constructive comments on an earlier draft of our work. We also thank Aotao
Xu, Emmy Liu, and Zhewei Sun for helping with the experimental design.
This research is supported by a NSERC Discovery Grant RGPIN-2018-05872349, a SSHRC Insight Grant \#435190272, and an Ontario Early Researcher Award \#ER19-15-050 to YX.
\end{acknowledgments}

\appendix
\appendixsection{Design details of the neural network modules}
Recall that the listener's and the speaker's distributions of three probabilistic models are parametrized by deep neural networks. For discriminative models, we use the BERT transformer to compute hidden representations for each token of the input sequence, and pass them through a fully-connected layer (parametrized by a transformation matrix $W$ and a bias vector $b$) to obtain proper probability distributions:
\begin{align}
    \text{p}_l(I=(V,R)|U) &= \sigma(W_{l,V}\cdot f_\text{BERT}(U) + b_{l,V}) \cdot \sigma(W_{l,R}\cdot f_\text{BERT}(U) + b_{l,R}) \\
    \text{p}_s(U=(D,C)|I) &= \sigma(W_{s,D}\cdot f_\text{BERT}(I) + b_{s,D}) \cdot \sigma(W_{s,C}\cdot f_\text{BERT}(I) + b_{s,C})
\end{align}
Here $\sigma$ is the softmax function, and we assume that conditioned on the input sequence, each component of the output sequence can be generated independently.

Similar to the discriminative model, both modules in the partial generative model first map input sequence into a hidden semantic space, and then sample each token of the output sequence independently by computing a categorical distribution for each component:
\begin{align}
    \text{p}_l(I=(V,R)|U) &= \sigma(W_{l,V}\cdot f_\text{l}(U) + b_{l,V}) \cdot \sigma(W_{l,R}\cdot f_\text{l}(U) + b_{l,R}) \\
    \text{p}_s(U=(D,C)|I) &= \sigma(W_{s,D}\cdot f_\text{s}(I) + b_{s,D}) \cdot \sigma(W_{s,C}\cdot f_\text{s}(I) + b_{s,C})
\end{align}
For the full generative model, the listener would also sample frame elements $E$ during inference, and the speaker would also take frame elements $E$ as input when sampling denominal utterances during generation:
\begin{align}
    \text{p}_l(I,E|U) &= \sigma(W_{l,V}\cdot f_\text{l}(U) + b_{l,V}) \cdot \sigma(W_{l,R}\cdot f_\text{l}(U) + b_{l,R}) \cdot \sigma(W_{l,E}\cdot f_\text{l}(U) + b_{l,E})\\
    \text{p}_s(U|I,E) &= \sigma(W_{s,D}\cdot f_\text{s}(I,E) + b_{s,D}) \cdot \sigma(W_{s,C}\cdot f_\text{s}(I,E) + b_{s,C})
\end{align}

\appendixsection{Mathematical proofs for variational learning}
Here we show that the variational learning scheme described can achieve the following two goals simultaneously: 1) finding a probabilistic model that maximizes likelihood of all unsupervised denominal utterances, and 2) finding a pair of listener-speaker modules that induce the same joint distribution $\text{Pr}(U,I)$ over denominal utterances and their interpretations. We shall use the full generative model for the proof, although the results should also apply to the partial generative model.

Suppose we have an unsupervised set of denominal utterances without any paraphrases available. To compute the probability that our generative model would generate a particular utterance $U$, we need to consider each possible meaning $M$ that may be associated with it, and then sum up all joint probabilities $\text{p}_s(U,M)$ defined by the model:
\begin{align}
    \text{p}_s(U) = \sum_{M} \text{p}_s(U,M)
\end{align}
The log-likelihood $\mathcal{J}$ of all utterances therefore has the form: 
\begin{align}
    \mathcal{J} &= \sum_{U^{(i)}\in X_\text{u}}  \log [\sum_{M}p_\text{s}(U^{(i)}|M)p_0(M) ]\\
                &= \sum_{U^{(i)}\in X_\text{u}}  \log [\E_{M\sim p_\text{0}}[p_\text{s}(U^{(i)}|M)] ]
\end{align}
where we use $\E_{M\sim p_\text{0}}$ to denote the process of taking expectation over all possible meanings.
However, optimizing $\mathcal{J}$ directly would be difficult for most cases, and a common alternative is first finding a lower bound of $\mathcal{J}$, and then maximizing that bound -- this is where we introduce the listener into the learning process. In particular, by inserting a listener's posterior $p_l(M|U)$ as an approximation of speaker's belief in utterance meanings $p_s(M|U)$, we can re-write Equation (18) as:
\begin{align}
    \mathcal{J} &= \sum_{U^{(i)}\in X_\text{u}}  \log [\sum_{M}p_\text{s}(U^{(i)}|M)p_0(M) ]\\
                &= \sum_{U^{(i)}\in X_\text{u}}  \log [\sum_{M}\frac{p_\text{s}(U^{(i)}|M)p_0(M)}{p_l(M|U)}p_l(M|U) ]
\end{align}
where we divide the joint probability $p_s(U,M)$ with the listener's meaning likelihood $p_l(U|M)$ and multiply it back. Using Jensen's Inequality and the concavity of the log function, we can therefore derive a lower bound for $\mathcal{J}$ by replacing the log-of-sum in (20) with a sum-of-log term:
\begin{align}
    \mathcal{J} &= \sum_{U^{(i)}\in X_\text{u}}  \log [\sum_{M}\frac{p_\text{s}(U^{(i)}|M)p_0(M)}{p_l(M|U)}p_l(M|U) ] \\
                &\geq \sum_{U^{(i)}\in X_\text{u}}\sum_{M} p_l(M|U)\log \frac{p_\text{s}(U^{(i)}|M)p_0(M)}{p_l(M|U)} \\
                &= \sum_{U^{(i)}\in X_\text{u}} \sum_{M}[\log p_s(U|M) - \log \frac{p_l(M|U)}{p_0(M)}p_l(M|U)] \\
                &= \sum_{U^{(i)}\in X_\text{u}} \E_{M\sim p_\text{0}}[\log p_s(U|M)] - D[p_\text{l}(M|U)||p_0(M|\alpha)] \\
                &= \mathcal{U}
\end{align}
Therefore, the unsupervised loss defined in Equation (3) is a universal lower bound of $\mathcal{J}$. Ideally, $\mathcal{U}$ and $\mathcal{J}$ would converge after training. In this case, the log-of-sum term in (21) will equal to the sum-of-log in (22), implying that the fractional term $\frac{p_\text{s}(U^{(i)}|M)p_0(M)}{p_l(M|U)}$ inside the log becomes constant (i.e., independent of $M$):
\begin{align}
    \frac{p_\text{s}(U^{(i)}|M)p_0(M)}{p_l(M|U)} = c
\end{align}
Moving $p_l(M|U)$ to the right-hand side and integrating over $M$ we have:
\begin{align}
    p_\text{s}(U^{(i)}) = c
\end{align}
So $p_l$ becomes the following:
\begin{align}
    p_l(M|U) &= \frac{p_\text{s}(U^{(i)}|M)p_0(M)}{c} \\
             &= \frac{p_\text{s}(U^{(i)}|M)p_0(M)}{p_\text{s}(U^{(i)})} \\
             &= p_s(M|U)
\end{align}
via Bayes' rule. Therefore, by optimizing $\mathcal{U}$, we can not only maximize the log-likelihood $J$ of all denominal utterances, but also operationalize the idea of shared semantic knowledge by forcing the listener and speaker module to define the same joint utterance-meaning distribution. 

\appendixsection{Prediction of  denominal verb usages in English-speaking adults and children}
\renewcommand{\thetable}{\Alph{section}\arabic{table}}
\renewcommand\thefigure{\Alph{section}\arabic{figure}}

\begin{table}[h]
\caption{Model comparison on predicting human annotated English denominal utterances made by adults and children. Model accuracy is summarized by Kullback-Leibler (KL) divergence between posterior distributions $p_\text{comp}(V|U)$, $p_\text{prod}(D|I)$ and fine-grained empirical distributions of human-annotated ground-truth on DENOM-AMT dataset. A lower value in KL indicates better alignment between model distribution and empirical distribution. Standard errors are shown within the parentheses.}
\resizebox{\textwidth}{!}{
\begin{tabular}{@{}lllll@{}}
\toprule
\multicolumn{1}{c}{}      & \multicolumn{4}{c}{KL divergence ($\times 10^{-3}$)}                               \\
\multicolumn{1}{c}{Model} & \multicolumn{2}{c}{English adults} & \multicolumn{2}{c}{English children} \\
                          & Comprehension   & Production   & Comprehension   & Production   \\ \midrule
Full Generative           & 16.8 (2.4)      & 53.1 (4.6)   & 29.7 (3.0)     & 92.5 (1.4)    \\
Partial Generative        & 19.1 (1.7)      & 56.5 (5.5)   & 31.1 (3.5)     & 115.7 (1.4)   \\
Discriminative            & 34.7 (2.2)      & 103.1 (3.9)  & 30.6 (2.9)     & 104.6 (1.3)   \\
Frequency Baseline        & 44.7 (0)        & 133.2 (0)    & 44.7 (0)        & 133.2 (0)     \\ \bottomrule
\end{tabular}
}
\label{table:kl-divs-breakdown}
\end{table}
\newpage

\begin{figure}[h]
\includegraphics[width=\linewidth]{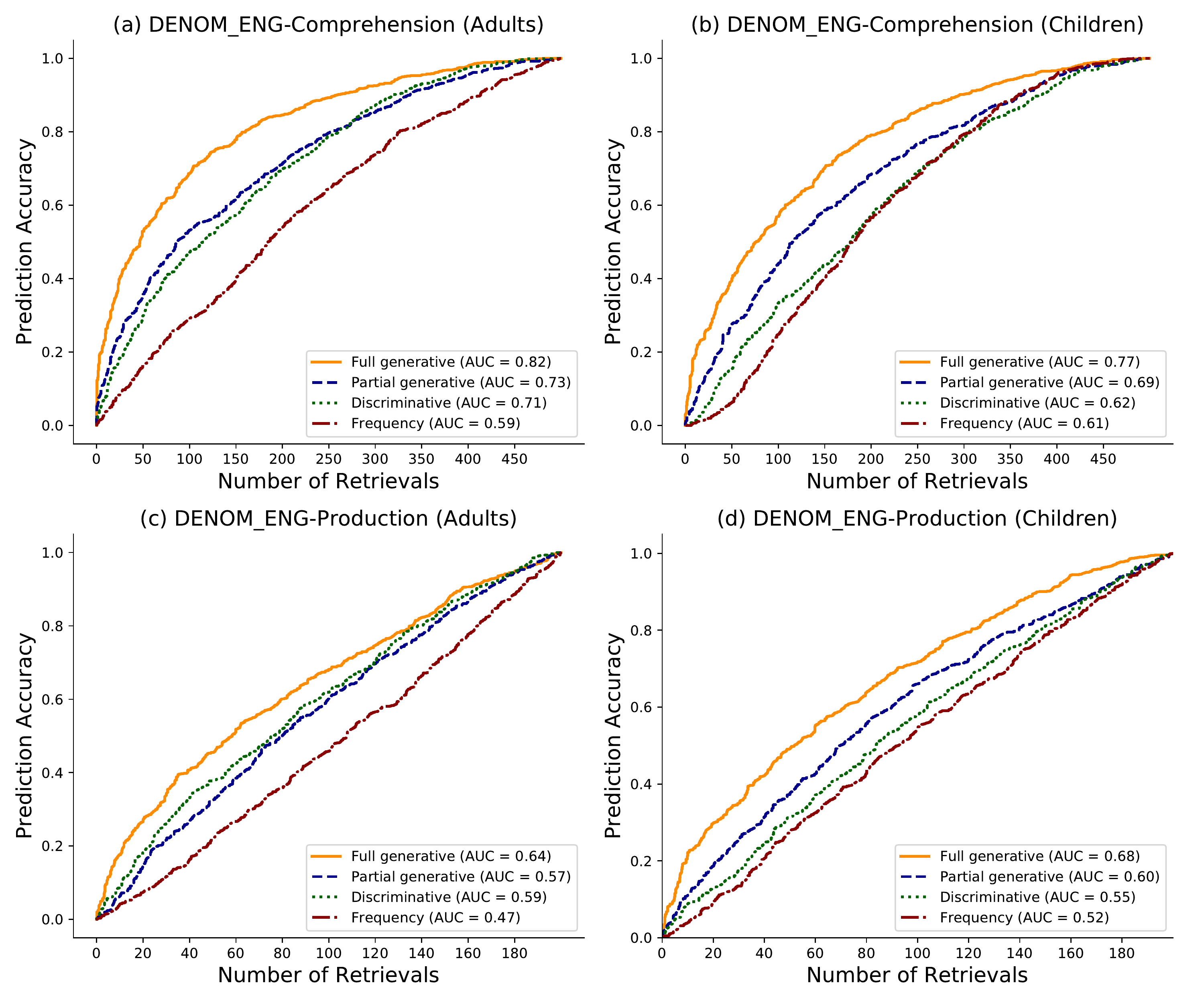}
\caption{A breakdown of model performance in English denominal verb comprehension and production, based on adults' and children's usage data. The left column summarizes the results from the 100 denominal utterances made by adults in DENOM-AMT dataset based on receiver operating characteristic (ROC) curves, and the right column summarizes similar results from the denominal utterance in DENOM-AMT made by children. ``Frequency'' refers to the frequency baseline model. Higher area-under-the-curve (AUC) score indicates better performance.}
\label{roc_curves_breakdown}
\end{figure}

\appendixsection{Calculation of KL divergence}
The Kullback-Leibler (KL) divergence $D_{KL}$ measures how one probability distribution is different from another reference distribution. For two discrete distributions $P, Q$,  their KL divergence is defined as:
\begin{equation}
    D_{KL}(P||Q) = \sum\limits_{x} P(x)\log (\frac{P(x)}{Q(x)})
\end{equation}
In our analysis, we compute the KL divergence between 1) the distribution of ground-truth responses in DENOM-AMT ($P$, proportional to the number of votes returned by human annotators), and 2) the model's output distribution of predicted word ($Q$). For instance, consider the following question of paraphrasing a denominal utterance: 

\begin{equation}
    \text{``I carpet the floor.''} \rightarrow \text{``I} \_\_\_ \text{the carpet on the floor.''}
\end{equation}
Suppose that there are 4 candidate paraphrase verbs with non-zero votes by the annotators and non-zero output probabilities returned by the full generative model:
\begin{table}[h]
\centering
\begin{tabular}{@{}lllll@{}}
\toprule
Candidate verb                                                                         & put  & drop & place & leave \\ \midrule
Vote by annotators                                                                     & 8    & 2    & 5     & 1     \\ \midrule
Induced empirical probability ($P$)                                                          & 0.5  & 0.13 & 0.31  & 0.06  \\ \midrule
\begin{tabular}[c]{@{}l@{}}Output probability by \\ full generative model  ($Q$)\end{tabular} & 0.41 & 0.08 & 0.16  & 0.01  \\ \bottomrule
\end{tabular}

\end{table}

The KL divergence can therefore be calculated as the following:
\begin{equation}
    D_{KL}(P||Q) = 0.5 * \log \frac{0.5}{0.41} + 0.13 * \log \frac{0.13}{0.08} + 0.31 * \log \frac{0.31}{0.16} + 0.02 * \log \frac{0.02}{0.01} = 0.38
\end{equation}

\starttwocolumn
\bibliography{wcc_main}

\end{document}